\documentclass[11pt]{article}

\usepackage[utf8]{inputenc}
\usepackage{amsmath,amssymb,amsthm}
\usepackage{array}
\usepackage{authblk}
\usepackage[english]{babel}
\usepackage{bm}
\usepackage{booktabs}
\usepackage{caption}
\usepackage{courier}
\usepackage{csquotes}
\usepackage[margin=1.0in]{geometry}
\usepackage{graphicx}
\usepackage{hyperref}
\usepackage{mathtools}
\usepackage[square,numbers]{natbib}
\usepackage{subcaption}
\usepackage{subfiles}

\graphicspath{ {images/} }

\hypersetup{
    colorlinks=false,
}

\title{\vspace{-1em}Deep learning in radiology: an overview of the concepts \\and a survey of the state of the art\\\vspace{0.2em}}
\date{\vspace{-2em}}

\author[1, 2, 3]{Maciej A. Mazurowski}
\author[1]{Mateusz Buda}
\author[1]{Ashirbani Saha}
\author[1, 4]{Mustafa R. Bashir}

\affil[ ]{\footnotesize }
\affil[1]{\footnotesize Department of Radiology, Duke University, Durham, NC}
\affil[2]{\footnotesize Department of Electrical and Computer Engineering, Duke University, Durham, NC}
\affil[3]{\footnotesize Duke Medical Physics Program, Duke University, Durham, NC}
\affil[4]{\footnotesize Center for Advanced Magnetic Resonance Development, Duke University, Durham, NC}

\begin{document}

\maketitle

\begin{abstract}

Deep learning is a branch of artificial intelligence where networks of simple interconnected units are used to extract patterns from data in order to solve complex problems.
Deep learning algorithms have shown groundbreaking performance in a variety of sophisticated tasks, especially those related to images.
They have often matched or exceeded human performance.
Since the medical field of radiology mostly relies on extracting useful information from images, it is a very natural application area for deep learning, and research in this area has rapidly grown in recent years.
In this article, we review the clinical reality of radiology and discuss the opportunities for application of deep learning algorithms.
We also introduce basic concepts of deep learning including convolutional neural networks.
Then, we present a survey of the research in deep learning applied to radiology.
We organize the studies by the types of specific tasks that they attempt to solve and review the broad range of utilized deep learning algorithms.
Finally, we briefly discuss opportunities and challenges for incorporating deep learning in the radiology practice of the future.


\end{abstract}

\section{Introduction}
\label{sec:introduction}

The field of deep learning encompasses a group of artificial intelligence methods which employ a large number of simple interconnected units to perform complicated tasks.
Deep learning algorithms, rather than using a set of pre-programmed instructions, are capable of learning from large amounts of data.
The tasks solved by these algorithms include localizing and classifying objects in images, understanding language, playing games, and many others.
While the flagship of deep learning, convolutional neural networks, were first introduced decades ago, it is only the last 5 years that showed an astonishing success of these algorithms elevating their status from an interesting but mostly impractical idea to the go-to algorithm in artificial intelligence.
In recent years, not only have deep learning algorithms been able to surpass performance of other methods in artificial intelligence~\cite{krizhevsky2012imagenet} but in some tasks, they have shown performance superior to humans~\cite{Dodge2017, he2015delving, rajpurkar2017chexnet}.

\begin{figure}[!ht]
    \centering
    \includegraphics[width=0.7\linewidth]{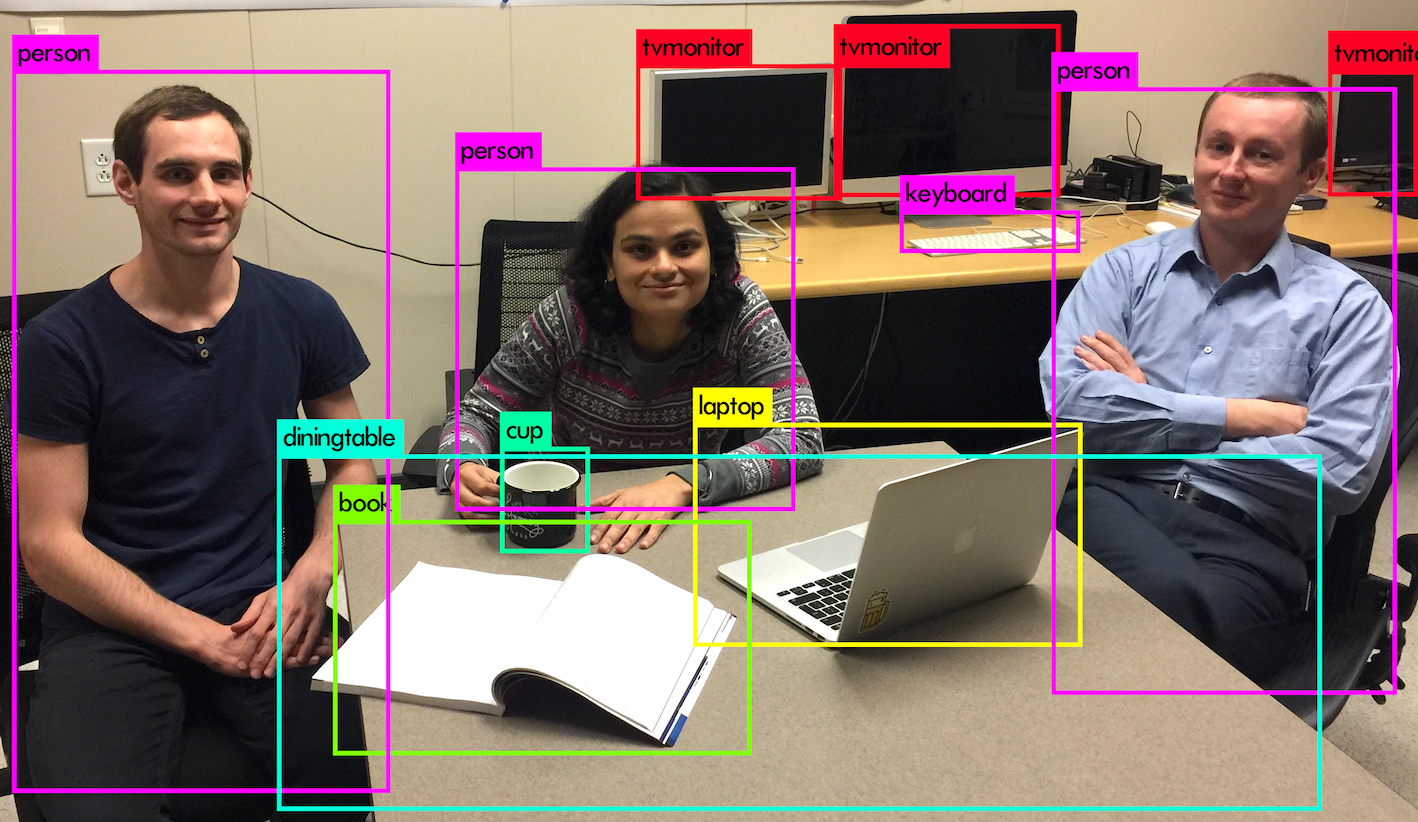}
    \caption{A picture of three of the authors of this article (Buda, Saha, Mazurowski) with detection labels generated by the YOLO detection network.}
    \label{fig:yolo}
\end{figure}

Arguably, the most well-known achievement of deep learning to date is its performance in the ImageNet competition.
ImageNet is a database of more than 14,000,000 annotated natural images containing real world objects such as cars, animals, and buildings (image-net.org).
One of the goals of the competition, started in 2009, is to assign each image to one of 1\,000 predefined categories.
When a deep learning-based algorithm first appeared in the competition in 2012, it dramatically improved the error rate from 0.258 in the previous year (image-net.org/challenges/LSVRC/2011/results) to 0.153 (image-net.org/challenges/LSVRC/2012/results).
The performance of deep learning algorithms for image classification has been improving since then and is now considered to be comparable to or better than human performance.
Other areas relevant to the topic of this article, where deep learning algorithms have seen impressive results, is in the automatic generation of sophisticated captions for images that consist of full sentences~\cite{Karpathy2015} as well as localization and outlining of objects in images~\cite{lin2017focal, Ren2015}.
To illustrate the capacity of current detection network, Figure~\ref{fig:yolo} shows the result of a deep neural network approach applied to detect objects in an image.

There are likely three reasons for the recent success of deep learning algorithms: availability of data, increased processing power, and rapid development of algorithms.
These are highly connected: availability of large datasets of images and computing power made it possible to demonstrate the strength of the basic concepts of deep learning and, motivated the development of further datasets and algorithms.
Increasing ease in applying algorithms and affordable graphical processing units have allowed for larger scientific and technical communities to get involved, and develop even more powerful algorithms which further advanced the field.

As the primary strength of deep learning has been in image analysis, the potential applications in radiology have become very quickly apparent.
The development of algorithms for radiology has shown some inertia due to the time needed for acquisition of the appropriate expertise in the medical imaging community as well as limited availability of large medical imaging datasets.
However, the last 2-3 years have seen remarkable productivity in the field.
It is now well recognized by both researchers and clinicians that deep learning will play a significant role in radiology.

In this paper, we begin with a general overview of radiology as the application domain and consider where deep learning could have the most significant impact.
Then, we introduce the general concepts of deep learning.
This is followed by an overview of the recent work in the field.
The article closes with remarks regarding the future of deep learning in radiology.

\section{The practice of radiology}
\label{sec:practice}

Radiology is a branch of medicine that focuses on using medical images for detection, diagnosis, and characterization of disease (diagnostic radiology) as well as guiding procedural interventions (interventional radiology).
In the United States, a typical radiologist undergoes 14-15 years of education after high school including 4 years of college, 4 years of medical school, 1 year of internship, 4 years of radiology residency, and generally 1-2 years of fellowship training.
While medical image interpretation work is centered in radiology practices and academic departments, some of it is also performed within other branches of medicine including cardiology, orthopedics, and surgery.

In diagnostic radiology, personal interaction between a radiologist with patients and other physicians is often limited.
The primary duty of a radiologist (particularly outside of an academic institution) is to view an image delivered to his/her reading station and generate a written report of findings.
This well-structured and isolated nature of the radiologist’s work makes is a particularly attractive application of artificial intelligence algorithms.

Deep learning techniques (and artificial intelligence algorithms in general) have a tremendous potential to influence the practice of radiology.
Unlike most other facets of medicine, all or nearly all of the primary data utilized in imaging is digital, lending itself to analysis by artificial intelligence algorithms.

In this section, we describe some of the primary challenges that a radiologist faces in his/her daily diagnostic radiology practice and briefly point to opportunities for deep learning to address them.
While this is not intended to be an exhaustive description of every tasks that radiologists perform in their practice, it reflects majority of diagnostic radiology practice.
We conclude this section with a description of some medical image interpretation tasks that are currently not performed by radiologists, but could be incorporated in radiology practice using deep learning.

\subsection{Disease Detection}
One of the most challenging tasks in the interpretation of imaging is the rapid differentiation of abnormalities from normal background anatomy.
For example, in the interpretation of mammography, each radiograph contains thousands of individual focal densities, regional densities, and geometric points and lines that must be interpreted to detect a small number of suspicious or abnormal findings.
In most cases, the entire mammogram should be interpreted as normal or negative, adding further complexity to the interpretive task.

In order to be useful, a computer algorithm does not have to detect all objects of interest (e.g., abnormalities) and be perfectly specific (i.e., not mark any normal locations).
For example, in screening mammography, approximately 80\% of screening mammograms should be read as negative according to the ACR BI-RADS guideline, and of the 20\% of examinations that trigger additional evaluation, many will ultimately be categorized as negative or benign~\cite{Ghate2005}.
An algorithm that could successfully categorize even half of screening mammograms as definitely negative would dramatically reduce the effort required to interpret a large batch of examinations.

\subsection{Disease Diagnosis and Management}
Once an abnormality has been detected, the often-complex task of determining a diagnosis and the disease management implications is undertaken.
For focal masses generically, a large number of features must be integrated in order to decide how to appropriately manage the finding.
These features can include size, location, attenuation or signal intensity, borders, heterogeneity, change over time, and others.
In some cases, simple criteria have been established and validated for the management of focal findings.
For example, most focal lesions in the kidney can be characterized as either simple or minimally complex cysts, which are almost uniformly benign.
On the other hand, most lesions in the kidney that are solid are considered to have high malignant potential.
Finally, a minority of focal kidney lesions is considered indeterminate and can be managed accordingly.

While for some types of abnormalities making the diagnostic and disease management decision follows straightforward guidelines, for other types of abnormalities, management algorithms are much more complex.
In the BI-RADS guideline for assessing focal lesions in the breast, a mass is categorized according to its shape (oval, round, or irregular), margin (circumscribed, obscured, microlobulated, indistinct, or spiculated), and its density (higher, equal to, or lower density than the glandular tissue, or fat-containing)~\cite{Ghate2005}.
Based on the constellation of features, the radiologist must then decide whether a mass is likely benign or requires follow-up or biopsy.
In the LI-RADS criteria for assessing focal liver lesions in patients at risk for developing hepatocellular carcinoma, five major features, and up to 21 ancillary features, are assessed to risk-stratify lesions and determine their management~\cite{Elsayes2017}.

Deep learning algorithms have the potential to assess a large number of features, even those previously not considered by radiologists, and arrive at a repeatable conclusion in a fraction of the time required for a human interpreter.
Perhaps most promisingly, these algorithms could be used to categorize large amounts of existing imaging data and correlate features with downstream health outcomes, a process that is currently extremely laborious and time-consuming when human interpretation is required.

\subsection{Workflow}
While detection, diagnosis, and characterization of disease receive the primary attention among algorithm developers, another important area where artificial intelligence could contribute is in facilitating the workflow of the radiologists while interpreting images.
With the widespread conversion from printed films to centralized Picture Archiving and Viewing Systems (PACS) as well as the availability of multi-planar, multi-contrast, and multi-phase imaging, radiologists have seen exponential growth in the size and complexity of image data to be analyzed.
Additionally, interpretations must often be rendered in the context of a multitude of prior examinations.
The simple task of finding and presenting these data is complex, and artificial intelligence systems may be well-suited for this role.

An example of a highly complex workflow is that for many cancer patients.
Such patients are not uncommonly afflicted with more than one primary tumor, metastatic disease to numerous sites, and may have undergone a variety of biopsies, locoregional therapies, and systemic therapies with varying results.
In the simplest scenario, interpretation of a follow-up imaging examination requires colocalization of all relevant sites of disease between the current and prior examinations.
Measurements of size are performed, and in some cases functional features, such as tumor perfusion or diffusion restriction, are assessed either subjectively or objectively.
Most radiology practices utilize imaging equipment of different types, generations, and often different vendors, thus simply identifying the appropriate image sets in prior examinations can be very challenging.
After the appropriate images have been identified, the radiologist must colocalize disease sites and attempt to obtain precise repeated measurements in order to ensure that the values obtained from the current and prior examinations can be compared.

Each of the above tasks is time-consuming and does not necessarily require the full skill of a radiologist.
However, standard PACS systems are not able to reliably present all of the above data for a variety of reasons, including the variability in labeling the types and components of imaging examinations, the variability in patient positioning and anatomy between examinations, the variability in modalities used to image the same portion of the anatomy, as well as other factors.
In principle, an artificial intelligence algorithm could assess a patient’s prior imaging, bring forward examinations that include the relevant body part(s), detect the image modality and contrast type, and determine the location of the area of interest within the relevant anatomy to reduce the radiologist’s effort in performing these relatively mundane tasks.

\subsection{Image interpretation tasks that radiologists do not perform but deep learning could}
In addition to performing tasks that are a part of the current radiological practice, computer algorithms could perform medical image interpretation tasks that radiologists do not perform on a regular basis.
The research toward this goal has been underway for some time, mostly using traditional machine learning and image processing algorithms.
One example is radiogenomics~\cite{mazurowski2015radiogenomics}, which aims to find relationships between imaging features of tumors and their genomic characteristics.
Examples can be found in breast cancer~\cite{Mazurowski2014}, glioblastoma~\cite{Gutman2013}, low grade glioma~\cite{mazurowski2017radiogenomics}, and kidney cancer~\cite{Karlo2014}.
Radiogenomics is not a part of the typical clinical practice of a radiologist.
Another example is prediction of outcomes of cancer patients with applications in glioblastoma~\cite{Gutman2013, Mazurowski2013}, lower grade glioma,~\cite{mazurowski2017radiogenomics}, and breast cancer~\cite{Mazurowski2015Recurrence}.
While imaging features have a potential to be informative of patient outcomes, very few are currently used to guide oncological treatment.
Deep learning could facilitate the process of incorporating more of the information available from imaging into the oncology practice.

\section{An introduction to deep learning}
\label{sec:deeplearning}

\subsection{Terminology}
To understand deep learning, it is helpful to first understand the related concepts of artificial intelligence and machine learning.
Artificial intelligence is a set of computer algorithms that are able to perform complicated tasks or tasks that require intelligence when conducted by humans.
Machine learning is a subset of artificial intelligence algorithms which, to perform these complicated tasks, are able to learn from provided data and do not require pre-defined rules of reasoning.
The field of machine learning is very diverse and has already had notable applications in medical imaging~\cite{Erickson2017}.
Deep learning is a sub-discipline of machine learning that relies on networks of simple interconnected units.
In deep learning models, these units are connected to form multiple layers that are capable of generating increasingly high level representations of the provided input (e.g., images).
Below, in order to explain the architecture of deep learning models, we introduce the artificial neural network in general and one specific type: the convolutional neural network.
Then, we detail the "learning" process of these networks, which is the process of incorporating the patterns extracted from data into deep neural networks.

\subsection{Artificial Neural Networks}
Artificial neural networks (ANNs) are machine learning models based on basic concepts dating as far as 1940s, significant development in 1970s and 1980s and a period notable popularity in 1990s and 2000s followed by a period of being overshadowed by other machine learning algorithms.
ANNs consist of a multitude of interconnected processing units, called neurons, usually organized in layers.
A traditional ANN typically used in the practice of machine learning contains 2 to 3 layers of neurons.
Each neuron performs a very simple operation.
While many neuron models were proposed, a typical neuron simply multiplies each input by a certain weight, then adds all the products for all the inputs and applies a simple nondecreasing function at the end.
Even though each neuron performs a very rudimentary calculation, the interconnected nature of the network allows for the performance of very sophisticated calculations and implementation of very complicated functions.

\subsection{Convolutional Neural Networks}
Deep neural networks are a special type of an ANN.
The most common type of a deep neural network is a deep convolutional neural network (CNN).
A deep convolutional neural network, while inheriting the properties of a generic ANN, has also its own specific features.
First, it is deep.
A typical number of layers is 10\nobreakdash-30 but in extreme cases it could exceed 1\,000.
Second, the neurons are connected such that multiple neurons share weights.
This effectively allows the network to perform convolutions (or template matching) of the input image with the filters (defined by the weights) within the CNN.
Other special feature of CNNs is that between some layers, they perform pooling which makes the network invariant to small shifts of the images.
Finally, CNNs typically use a different activation function of the neurons as compared to traditional ANNs.

\begin{figure}[!ht]
    \centering
    \includegraphics[width=0.7\linewidth]{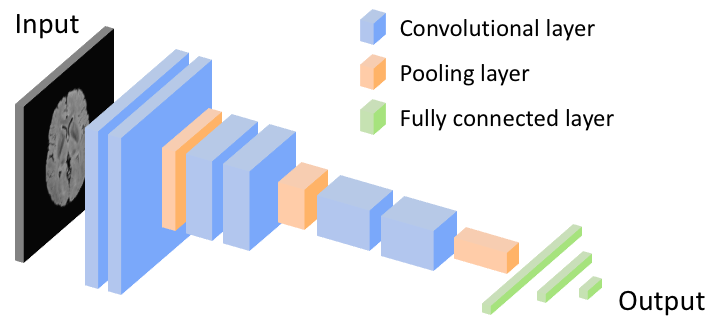}
    \caption{A diagram illustrating a typical architecture of a convolutional neural network.}
    \label{fig:cnn}
\end{figure}

Figure~\ref{fig:cnn} shows an example of a small architecture for a typical CNN.
One can see that the first layers are the convolutional ones which serve the role of generating useful features for classification.
Those layers can be thought of as implementing image filters, ranging from simple filters that match edges to those that eventually match much more complicated shapes such as eyes, or tumors.
Further from the network input are so called fully connected layers (similar to traditional ANNs) which utilize the features extracted by the convolutional layers to generate a decision (e.g., assign a label).
A variety of deep learning architectures have been proposed, often driven by characteristics of the task at hand (e.g., fully convolutional neural networks for image segmentation).
Some of these are described in more detail in the section of this paper that reviews the current state of the art.

\subsection{The learning process in convolutional neural networks}
Above, we described general characteristics of traditional neural networks and deep learning’s flagship: the convolutional neural network.
Next, we will explore how to make those networks perform useful tasks.
This is accomplished in the process referred to as learning or training.
The learning process of a convolutional neural network simply consists of changing the weights of the individual neurons in response to the provided data.
In the most popular type of learning process, called supervised learning, a training example contains an object of interest (e.g., an image of a tumor) and a label (e.g., the tumor’s pathology: benign or malignant).
In our example, the image is presented to the network’s input, and the calculation is carried out within the network to produce a prediction based on the current weights of the network.
Then, the network’s prediction is compared to the actual label of the object and an error is calculated.
This error is then propagated through the network to change the values of the network's weights such that the next time the network analyzes this example, the error decreases.
In practice, the adaptation of the weights is performed after a group of examples (a batch) are presented to the network.
This process is called error backpropagation or stochastic gradient descent.
Various modifications of stochastic gradient descent algorithm have been developed~\cite{ruder2016overview}.
In principle, this iterative process consists of calculations of error between the output of the model and the desired output and adjusting the weights in the direction where the error decreases.

\begin{figure}[!ht]
    \centering
    \includegraphics[width=0.7\linewidth]{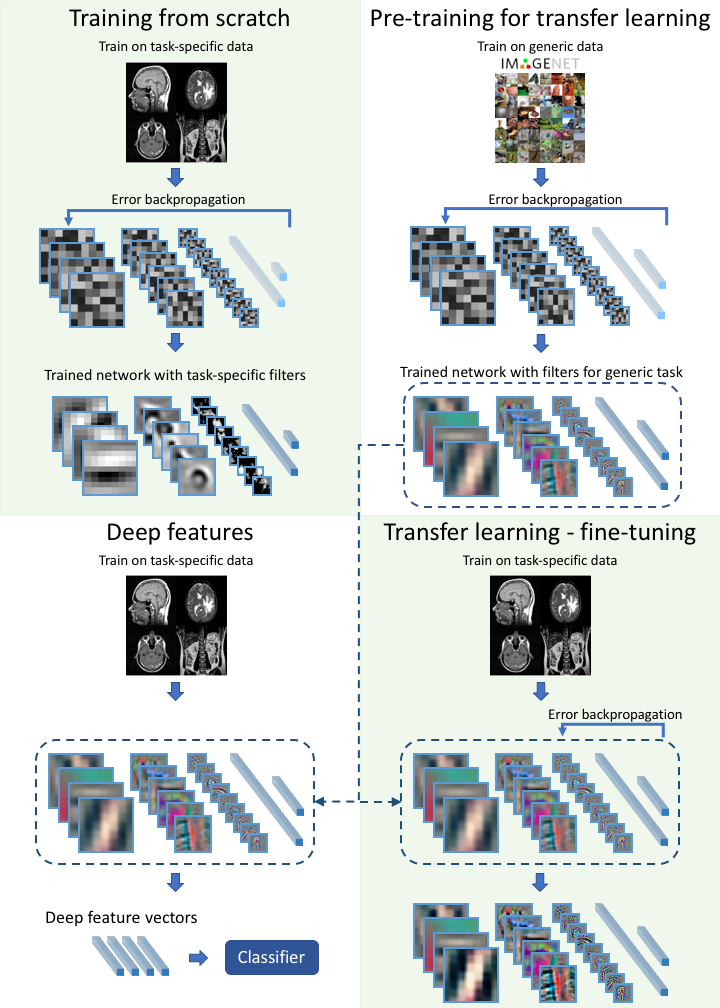}
    \caption{An illustration of different ways of training in deep neural networks.}
    \label{fig:training}
\end{figure}

The most straightforward way of training is to start with a random set of weights and train using available data specific to the problem being solved (training from scratch).
However, given the large number of parameters (weights) in a network, often above 10 million, and a limited amount of training data (common in medical imaging), a network may overfit to the available data, resulting in poor performance on test data.
Two training methods have been developed to address this issue: transfer learning~\cite{yosinski2014transferable} and off-the-shelf features (a.k.a. deep features)~\cite{sharif2014cnn}.
A diagram comparing training from scratch with transfer learning and off-the-shelf deep features is shown in Figure~\ref{fig:training}.

In the transfer learning approach, the network is first trained using a different dataset, for example an ImageNet collection.
Then, the network is "fine-tuned" through additional training with data specific to the problem to be addressed.
The idea behind this approach is that solving different visual tasks shares a certain level of processing such as recognition of edges or simple shapes.
This approach has been shown successful in, for example, prediction of survival time from brain MRI in patients with glioblastoma tumor~\cite{ahmed2017fine} or in skin lesion classification~\cite{Esteva}.
Another approach that addresses the issue of limited training data is the deep "off-the-shelf" features approach which uses convolutional neural networks which have been trained on a different dataset to extract features from the images.
This is done by extracting outputs of layers prior to the network's final layer.
Those layers typically have hundreds or thousands of outputs.
Then, these outputs are used as inputs to "traditional" classifiers such as linear discriminant analysis, support vector machines, or decision trees.
This is similar to transfer learning (and is sometimes considered a part of transfer learning) with the difference being that the last layers of a CNN are replaced by a traditional classifier and the early layers are not additionally trained.

\subsection{Deep learning vs "traditional" machine learning}
\begin{figure}[!ht]
    \centering
    \includegraphics[width=0.66\linewidth]{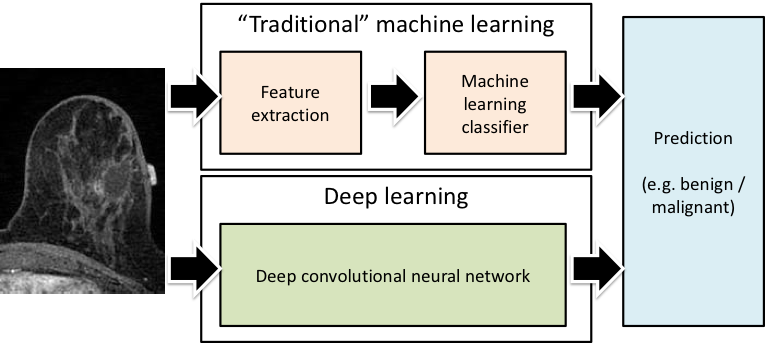}
    \caption{An illustration of difference between “traditional” machine learning and deep learning.}
    \label{fig:comparison}
\end{figure}

Increasingly often we hear a distinction between deep learning and "traditional" machine learning (see Figure~\ref{fig:comparison}).
The difference is very important, particularly in the context of medical imaging.
In traditional machine learning, the typical first step is feature extraction.
This means that to classify an object, one must decide which characteristics of an object will be important and implement algorithms that are able to capture these characteristics.
A number of sophisticated algorithms in the field of computer vision have been proposed for this purpose and a variety of size, shape, texture, and other features were extracted.
This process is to a large extent arbitrary since the machine learning researcher or practitioner often must guess which features will be of use for a particular task and runs the risk of including useless and redundant features and, more importantly, not including truly useful features.
In deep learning, the process of feature extraction and decision making are merged and trainable, and therefore no choices need to be made regarding which features should be extracted; this is decided by the network in the training process.
However, the cost of allowing the neural network to select its own features is a requirement for much larger training data sets.

\section{Deep learning in radiology: state of the art}
\label{sec:sota}

In this section, we give an overview of applications of deep learning in radiology.
We organized this section by the tasks that the deep learning algorithms perform.
Within each subsection, we describe different methods applied, and when possible, we systematically discuss the evolution of these methods in the recent years.

\subsection{Classification}
In a classification task, an object is assigned to one of the predefined classes.
A number of different classification tasks can be found in the domain of radiology such as: classification of an image or an examination to determine the presence or an absence of an abnormality; classification of abnormalities as benign or malignant; classification of cancerous lesions according to their histopathological and genomic features; prognostication; and classification for the purpose of organization radiological data. 

Deep learning is becoming the methodology of choice for classifying radiological data.
The majority of the available deep learning classifiers use convolutional neural networks with a varying number of convolutional layers followed by fully connected layers.
The availability of radiological data is limited as compared to the natural image datasets which drove the development of deep learning techniques in the last 5 years.
Therefore, many applications of deep learning in medical image classification have resorted to techniques meant to alleviate this issue: off-the-shelf features and transfer learning~\cite{tajbakhsh2016convolutional} discussed in the previous section of this article.
Off-the-shelf features have performed well in a variety of domains~\cite{sharif2014cnn}, and this technique has been successfully applied to medical imaging~\cite{Antropova2017, Paul2016}.
In~\cite{Antropova2017}, the authors combined the deep off-the shelf features extracted from a pre-trained VGG19 network with hand-crafted features for determining malignancy of breast lesions in mammography, ultrasound, and MRI.
In~\cite{Paul2016}, long-term and short term survival was predicted for patients with lung carcinoma.
The transfer learning strategy, which involves fine tuning of a network pre-trained on a different dataset, has been applied to a variety of tasks such as classification of prostate MR images to distinguish patients with prostate cancer from patients with benign prostate conditions~\cite{Wang2017a}, identification of CT images with pulmonary tuberculosis~\cite{Lakhani2017}, and classification of radiographs to identify hip osteoarthritis~\cite{Xue2017}.
Most of the studies which apply the transfer learning strategy replace and retrain the deepest layer of a network, whereas shallow layers are fixed after the initial training.
A variation of the transfer learning strategy combines fine-tuning and deep features approach.
It fine-tunes a pre-trained network on a new dataset to obtain more task-specific deep feature representations.
An example of this is the study~\cite{Chi2017}, which performed ultrasound imaging-based thyroid nodule classification using features extracted from a fine-tuned pre-trained GoogLeNet.
An ensemble of fine-tuned CNN classifiers was shown to predict radiological image modality in the study~\cite{Kumar2017}.
A comparison of approaches using deep features and transfer learning with fine tuning was shown in the study~\cite{Zhu2017a} identifying radiogenomic relationships in breast cancer MR images and in~\cite{Zhu2017} for predicting the upstaging of ductal carcinoma in situ to invasive breast cancer from breast cancer MR images.
In both of these problems, deep features performed better than transfer learning with the fine tuning approach.
However, both of these studies faced the issue of a small size of the training set.

When sufficient data are available, an entire deep neural network can be trained from a random initialization (training from scratch).
The size of the network to be trained depends on task and dataset characteristics.
However, the commonly used architecture in medical imaging is based on AlexNet~\cite{krizhevsky2012imagenet} and VGG~\cite{simonyan2014very} with modifications that have fewer layers and weights.
Examples of training from scratch can also be found in various studies such as: assessing for the presence of Alzheimer’s disease based on brain MRI using deep learning~\cite{li2017deep, Suk2017}, glioma grading in MRI~\cite{Khawaldeh}, and disease staging and prognosis in chest CT of smokers~\cite{Gonzalez2017}.
Recent advances in the design of CNN architectures has made networks easier to train and more efficient.
They have more layers and perform better while having fewer trainable parameters~\cite{canziani2016analysis} which reduces the likelihood of overtraining.
The most notable examples include Residual Networks (ResNets)~\cite{he2016deep} and the Inception architecture~\cite{szegedy2017inception, szegedy2016rethinking}.
A shift to these more powerful networks has also taken place in applications of deep learning to radiology both for transfer learning and training from scratch.
Three different ResNets were used to predict methylation of the O6-methylguanine methyltransferase gene status from pre-surgical brain tumor MRI~\cite{Korfiatis2017}.
In~\cite{Kim2017}, the InceptionV3 network was fine-tuned and served as a feature extractor instead of previously used GoogLeNet.
In another work using chest X-ray images~\cite{rajpurkar2017chexnet}, the authors fine-tuned a DenseNet with 121 layers for the classification of miscellaneous pathologies, achieving radiologist-level classification performance for identifying pneumonia.

In another approach, auto-encoder (AE)~\cite{hinton2006reducing} or stacked auto-encoder (SAE)~\cite{bengio2007greedy, poultney2007efficient} networks have been trained from scratch, layer by layer in unsupervised way.
A stacked denoising auto-encoder with backpropagation was used in~\cite{Ortiz2017} to determine the presence of Alzheimer’s disease.
AEs and SAEs can also be used to extract feature representations (similarly to the deep features approach) from hidden layers for further classification.
Such feature representation has been used in the classification of lung nodules into benign and malignant classes in CT~\cite{Kumar2015}, and in the identification of multiple sclerosis lesions in MRI~\cite{Yoo2018}.

Apart from the classification of radiological images, analysis of radiological text reports plays a significant role~\cite{Wang2012}.
The most prominent approach in this type of classification is deep learning-based natural language processing (NLP)~\cite{Kim2014}, which is based on the seminal work for obtaining vector representation of phrases using an unsupervised neural model~\cite{Mikolov2013}.
An example of application of this architecture can be found in~\cite{Chen2017a} where the authors classified CT radiology reports as representing presence or absence of pulmonary embolism (PE), as well as type (chronic or acute)  and location (central or subsegmental) of PE when present.
They showed an improvement as compared to a non-deep learning algorithm.
The same architecture was used in~\cite{Shin2017} for classifying head CT reports of ICU patients with altered mental status as having different degrees of severity according to each of five criteria: severity of study, acute intracranial bleed, acute mass effect, acute stroke, acute hydrocephalus.
Radiology reports using the International Coding of Diseases (ICD) were auto-encoded in~\cite{Karimi2017} using a publicly available dataset.
A third application of the same architecture can be found in~\cite{Karimi2017} where radiology reports were classified according to the International Coding of Diseases9 (ICD9) using a publicly available dataset.

\subsection{Segmentation}
In an image segmentation task, an image is divided into different regions in order to separate distinct parts or objects.
In radiology, the common applications are segmentation of organs, substructures, or lesions, often as a preprocessing step for feature extraction and classification~\cite{li2017deep, akkus2017predicting}.
Below, we discuss different types of deep learning approaches used in segmentation tasks in a variety of radiological images.

The most straightforward and still widely used method for image segmentation is classification of individual pixels based on small image patches (both 2-dimensional and 3-dimensional) extracted around the classified pixel.
This approach has found usage in different types of segmentation tasks in MRI, for example brain tumor segmentation in~\cite{Havaei2015, hussain2017brain, milletari2017hough}, white matter segmentation in multiple sclerosis patients~\cite{valverde2017improving}, segmentation of 25 different structures in brain~\cite{wachinger2017deepnat}, and for rectal cancer segmentation in pelvis MRI~\cite{trebeschi2017deep}.
It allows for using the same network architectures and solutions that are known to work well for classification, however, there are some shortcomings of this method.
The primary issue is that it is computationally inefficient, since it processes overlapping parts of images multiple times.
Another drawback is that each pixel is segmented based on a limited-size context window and ignores the wider context.
In some cases, a piece of global information, e.g. pixel location or relative position to other image parts, may be needed to correctly assign its label.

One approach that addresses the shortcomings of the pixel-based segmentation is a fully convolutional neural network (fCNN)~\cite{long2015fully}.
Networks of this type process the entire image (or large portions of it) at the same time and output a 2-dimensional map of labels (i.e., a segmentation map) instead of a label for a single pixel.
Example architectures that were successfully used in both natural images and radiology applications are encoder-decoder architectures such as U-Net~\cite{christ2017automatic, ronneberger2015u, salehi2017real} or Fully Convolutional DenseNet~\cite{jegou2017one, chenmri, li2017h}.
Various adjustments to these types of architectures have been developed that mainly focus on connections between the encoder and decoder parts of the networks, called skip connections. Applications of fCNNs in radiology include prostate gland segmentation in MRI~\cite{clark2017fully}, segmentation of multiple sclerosis lesions and gliomas in MRI~\cite{mckinley2016nabla}, and ultrasound-based nerve segmentation~\cite{zhang2017image}.
Moreover, loss functions have been explored that account for class imbalance (differences in the number of examples in each class), which is typical in medical datasets, e.g. weighted cross entropy was used in~\cite{mehta2017m} for brain structure segmentation in MRI or Dice coefficient-based loss for brain tumor segmentation in MRI~\cite{sudre2017generalised}.

In order to segment 3-dimensional data, it is common to process data as 2-dimensional slices and then combine the 2-dimensional segmentation maps into a 3-dimensiaonal map since 3D fCNNs are significantly larger in terms of trainable parameters and as a result require significantly larger amounts of data.
Nevertheless, these obstacles can be overcome, and there are successful applications of 3D fCNNs in radiology, e.g. V-Net for prostate segmentation from MRI~\cite{milletari2016v} and 3D U-Net~\cite{cciccek20163d} for segmentation of the proximal femur in MRI~\cite{deniz2017segmentation} and tumor segmentation in multimodal brain MRI~\cite{shenmultimodal}.

Finally, a deep learning approach that has found some application in medical imaging segmentation is recurrent neural networks (RNNs).
In~\cite{yang2017fine}, the authors used a Boundary Completion RNN for prostate segmentation in ultrasound images.
Another notable application is in~\cite{poudel2016recurrent}, where the authors applied a recurrent fully convolutional neural network for left-ventricle segmentation in multi-slice cardiac MRI  to leverage inter-slice spatial dependencies.
Similarly, \cite{cai2017improving} used Long Short-Term Memory (LSTM)~\cite{hochreiter1997long} type of RNN trained end-to-end together with fCNN to take advantage of 3D contextual information for pancreas segmentation in CT and MR images.
In addition, they proposed a novel loss function that directly optimizes a widely used segmentation metric, the Jaccard Index~\cite{jaccard1912distribution}.

\subsection{Detection}
Detection is a task of localizing and pointing out (e.g., using a rectangular box) an object in an image.
In radiology, detection is often an important step in the diagnostic process which identifies an abnormality (such as a mass or a nodule), an organ, an anatomical structure, or a region of interest for further classification or segmentation~\cite{al2010improved, oliver2010review, rey2002automatic}.
Here, we discuss the common architectures used for various detection tasks in radiology along with example specific applications.

The most common approach to detection for 2-dimensional data is a 2-phase process that requires training of 2 models.
The first phase identifies all suspicious regions that may contain the object of interest.
The requirement for this phase is high sensitivity~\cite{roth2014new} and therefore it usually produces many false positives.
A typical deep learning approach for this phase is a regression network for bounding box coordinates based on architectures used for classification~\cite{erhan2014scalable, szegedy2014scalable}.
The second phase is simply classification of sub-images extracted in the previous step.
In some applications, only one of the two steps uses deep learning.
The classification step, when utilizing deep learning, is usually performed using transfer learning.
The models are often pre-trained using natural images, for example for thoraco-abdominal lymph node detection in~\cite{shin2016deep} and pulmonary embolism detection in CT pulmonary angiogram images~\cite{tajbakhsh2016convolutional}.
In other applications, the models are pre-trained using other medical imaging dataset to detect masses in digital breast tomosynthesis images~\cite{samala2016mass}.
The same network architectures can be used for the second phase as in a regular classification task (e.g., VGG~\cite{simonyan2014very}, GoogLeNet~\cite{szegedy2015going}, Inception~\cite{szegedy2017inception}, ResNet~\cite{he2016deep}) depending on the needs of a particular application.

While in the 2-phase detection process the models are trained separately for each phase, in the end-to-end approach one model encompassing both phases is trained.
An end-to-end architecture that has proved to be successful in object detection in natural images, and was recently applied to medical imaging, is the Faster Region-based Convolutional Neural Network (R-CNN)~\cite{Ren2015}.
It uses a CNN to obtain a feature map which is shared between region proposal network that outputs bounding box candidates, and a classification network which predicts the category of each candidate.
It was recently applied for intervertebral disc detection in X-ray images~\cite{sa2017intervertebral} and detection of colitis on CT images~\cite{liu2017detection}.
A domain specific modification that uses additional preprocessing before the region proposal step was used by~\cite{ben2017domain} for detection of architectural distortions in mammograms.

Another approach to detection is a single-phase detector that eliminates the first phase of region proposals.
Examples of popular methods that were first developed for detection in natural images and rely on this approach are You Only Look Once (YOLO)~\cite{redmon2016you}, Single Shot MultiBox Detector (SSD)~\cite{liu2016ssd} and RetinaNet~\cite{lin2017focal}.
In the context of radiology, a YOLO-based network called BC-DROID has been developed by~\cite{platania2017automated} for region of interest detection in breast mammograms.
SSD has been employed, for example in~\cite{cao2017breast}, for breast tumor detection in ultrasound images, outperforming other evaluated deep learning methods that were available at the time.
The authors of~\cite{li2017detection} applied the same network for detection of pulmonary lung nodules in CT images.

In the examples above, 2-dimensional data was used.
For 3-dimensional imaging volumes, common in medical imaging, results obtained from 2-dimensional processing are combined to produce the ultimate 3-dimensional bounding box.
As an example, in \cite{de20162d} the authors performed detection of 3D anatomy in chest CT images by processing data slice by slice in one direction.
Combining output from different planes was performed in several studies.
Most of the them~\cite{prasoon2013deep, roth2016deep, roth2016improving} used orthogonal planes of MRI and CT images performing detection in each direction separately.
The results can then be combined in different ways, e.g. by an algorithm based on output probabilities~\cite{de20162d} or using another machine learning method like random forest~\cite{li2017detection}.
An alternative method for 3D detection has been proposed for automatic detection of lymph nodes using CT images by concatenating coronal, sagittal and axial views as a single 3-channel image in~\cite{roth2014new}.

\subsection{Other Tasks in Radiology}
While the majority of the applications of deep learning in radiology have been in classification, segmentation, and detection, other medical imaging-related problems have found some solutions in deep learning.
Due the variety of those problems, there is no unifying methodological framework for these solutions.
Therefore below, we organize the examples according to the problem that they attempt to address.

\paragraph{Image Registration:}
In this task two or more images (often 3D volumes), typically of different types (e.g., T1-weighted and T2-weighted MRIs) must be spatially aligned such that the same location in each image represents the same physical location in the depicted organ.
Several approaches can be taken to address the problem.
In one approach, it is necessary to calculate similarity measures between image patches taken from the images of interest to register them.
The authors of~\cite{Simonovsky2016} used deep learning to learn a similarity measure from T1-T2 MRI image pairs of adult brain and tested it to register T1-T2 MRI interpatient images of the neonatal brain.
This similarity measure performed better than the standard measure, called mutual information, which is widely used in registration~\cite{Maes1997}.
In another deep learning-based approach to image registration, the deformation parameters between image pairs are directly learned using misaligned image pairs.
In~\cite{Liao2017}, the authors trained a CNN-based model to learn the sequence of movements that resulted in the misalignment of the image pairs of  CT and cone-beam CT examinations of the abdominal spine and heart.
In another study~\cite{Sokooti2017}, chest CT follow-up examinations were registered by training a CNN to predict three-dimensional displacement vector fields between the fixed and moving image pairs.
A CNN-based network was trained to correct respiratory motion in 3D abdominal MR images by predicting spatial transforms~\cite{Lv2017}.
All of these techniques are supervised regression techniques as they were trained using ground truth deformation information.
In another approach~\cite{DeVos2017}, which was unsupervised, a CNN was trained end-to-end to generate a spatial transformation which minimized dissimilarity between misaligned image pairs.

\paragraph{Image generation:}
Acquisition parameters of a radiological image strongly affect the visual quality and detail of the images obtained using the same modality.
First, we discuss the applications that synthesize images generated using different acquisition parameters within the same modality.
In~\cite{Bahrami2016}, 7T like images were generated from 3T MR images by training a CNN with patches centered around voxels in the 3T MR images.
Undersampled (in k-space) cardiac MRIs were reconstructed using a deep cascade of CNNs in~\cite{Schlemper2017}.
A real-time method to reconstruct compressed sensed MRI using GAN was proposed by~\cite{Yang2017}.
In another approach~\cite{Chartsias2017} in order to synthesize brain MRI images based on other MRI sequences in the same patient, convolutional encoders were built to generate a latent representation of images.
Then, based on that representation a sequence of interest was generated.
Reconstruction of "normal-dose" CT images from low-dose CT images (which are degraded in comparison to normal-dose images) has been performed using patch-by-patch mapping of low-dose images to high-dose images using a shallow CNN~\cite{Chen2017}.
In contrast, a deep CNN has been trained with low-dose abdominal CT images for reconstruction of normal-dose CT~\cite{Kang2017}.
In another study, CT images were reconstructed from a lower number of views using an U-Net inspired architecture~\cite{Jin2017}.

Deep learning has also been applied to synthesizing images of different modalities.
For example, CT images have been generated using MRIs by adopting an FCN to learn an end-to-end  non-linear mapping between pelvic CTs and MRIs~\cite{Nie2016}.
Synthetic CT images of brain were generated from one T1-weighted MRI sequence in~\cite{Han2017}.
In another application to aid a classification framework for Alzheimer’s disease diagnosis with missing PET scans, PET patterns were predicted from MRI using CNN~\cite{Li2014}.

\paragraph{Image enhancement:}
Image enhancement aims to improve different characteristics of the image such as resolution, signal-to-noise-ratio, and necessary anatomical structures (by suppressing unnecessary information) through various approaches such as super-resolution and denoising.

Super-resolution of images is important specifically in cardiac and lung imaging.
Three dimensional  near–isotropic  cardiac and lung images often require long scan times in comparison to the time the subject can hold his or her breath.
Thus, multiple 2D slices are acquired instead and the super-resolution methodology is applied to improve the resolution of the images.
An example of using deep learning in super-resolution in cardiac MRI can be found in~\cite{Oktay2016}, where the authors developed different models for single image super-resolution and for generating high resolution three-dimensional image volumes from two-dimensional image stacks.
In another study using CT, a single image super-resolution approach based on CNN was applied in a publicly available chest CT image dataset to generate high-resolution CT images, which are preferred for interstitial lung disease detection~\cite{Umehara2017}.
In this study, upscaled bicubic-interpolated images were first passed through one convolutional layer to generate low-resolution features.
Then, a non-linear transformation of those features was mapped to generate high resolution image features for the reconstruction.  

An example of an application of deep learning in denoising can be found in~\cite{Benou2017} where the authors performed denoising of DCE-MRI images of a brain (for stroke and brain tumors) by training an ensemble of deep auto-encoders using synthesized data.
Removal of Rician noise in MR images using a deep convolutional neural network aided with residual learning was performed in~\cite{Jiang2017}.

In an attempt to enhance the visual details of lung structure in chest radiographs, the effect of bone structures (ribs and clavicles) were suppressed.
Bone structure has been estimated by conditional random field based fusion of the outputs of a cascaded architecture of CNNs at multiple scales~\cite{Yang2017a}.
Metal artifacts (caused by prosthetics, dental procedures etc.) have also been suppressed by using a trained CNN model to generate metal-free images using CT~\cite{Zhang2017}.

\paragraph{Content-based image retrieval:}
In the most typical version of this task, the algorithm, given a query image, finds the most similar images in a given database.
To accomplish this task, in~\cite{Qayyum2017}, a deep CNN was first trained to distinguish between different organs.
Then, features from the three fully connected layers in the network were extracted for the images in the set from which the images were retrieved (evaluation dataset).
The same features were then extracted from the query image and compared with those of the evaluation dataset to retrieve the image.
In another study, a method was developed to retrieve, arrange, and learn the relationships between lesions in CT images~\cite{Yan2017}.

\paragraph{Objective image quality assessment:}
Objective quality assessment measures of medical images aim to classify an image to be of satisfactory or unsatisfactory quality for the subsequent tasks.
Objective quality measures of medical images are important to improve diagnosis and aid in better treatment~\cite{Chow2016}.
Image quality of fetal ultrasound was predicted using CNN in a recent study~\cite{Wu2017}.
Another study \cite{Abdi2017}, attempted to reduce the data acquisition variability in echocardiograms using a CNN trained on the quality scores assigned by an expert radiologist.
Using a simple CNN architecture, T2-weighted liver MR images were classified as diagnostic or non-diagnostic quality by CNN in~\cite{Esses2017}.

\section{Future of deep learning in radiology}
\label{sec:future}

There is a general agreement that deep learning will play a role in the future practice of radiology.
Some predict that it will conduct mundane tasks leaving radiologists with more time to focus on intellectually demanding challenges.
Some believe that radiologists and deep learning algorithms will work hand-in-hand to deliver performance superior to either of them alone.
Finally, some predict that deep learning algorithms will replace radiologists (at least in their image interpretation capacity) altogether.

Incorporation of deep learning in radiology will be associated with multiple challenges.
First, and currently foremost, is the technological challenge.
While deep learning has shown extraordinary promise in other image-related tasks, the results in radiology are far from showing that deep learning algorithms will replace a radiologist in the entire scope of their diagnostic work.
Some recent studies~\cite{Gale2017, Grewal2017, Jamaludin2017, Kooi2017, Larson2017, Merkow2017, Olczak2017, rajpurkar2017chexnet} indicate performance of these algorithms comparable to expert humans, but these results are only applicable to a very small minority of the tasks that radiologists perform.
This is likely to change in upcoming years given the rapid progress in implementing the deep learning algorithms in the realm of radiology.

Implementation of deep learning in radiology practice also poses legal and ethical challenges.
Primarily: who will be responsible for the mistakes that a computer will make?
While this is a difficult question, similar questions have been posed and resolved when other technologies were introduced including elevators and cars.
Since artificial intelligence penetrates various areas of human activity, questions of this type will likely be studied and answers proposed in the coming years.

Other challenges will include patient acceptance or non-acceptance of a radiologist's not being involved in the process of interpreting their images (regardless of the performance) as well as regulatory issues.
Finally, an important practical issue is how to incorporate deep learning algorithms into the radiology workflow in order to improve, rather than disrupt the radiology practice.

\section{Conclusion}
\label{sec:Conclusion}

In summary, in this paper we have discussed the principles of deep learning as well as the current practice of radiology to elucidate how these new algorithms can be incorporated into radiology workflow. We have discussed the progress and state of art in the field.
Finally, we have discussed some challenges and questions related to implementation of deep learning in the current practice of medicine.
All signs show that deep learning will play a significant role in radiology.
The next 5 years will be a very exciting time in the field that may see many questions stated in this article answered through a collaboration of machine learning scientists and radiologists.

\vspace{2em}

\noindent
\textbf{Acknowledgments:} The authors would like to acknowledge funding from the National Institutes of Biomedical Imaging and Bioengineering grant 5 R01 EB021360.
The authors would like to thank Gemini Janas for reviewing and editing this article.

\bibliography{references}

\begin{thebibliography}{139}
\providecommand{\natexlab}[1]{#1}
\providecommand{\url}[1]{\texttt{#1}}
\expandafter\ifx\csname urlstyle\endcsname\relax
  \providecommand{\doi}[1]{doi: #1}\else
  \providecommand{\doi}{doi: \begingroup \urlstyle{rm}\Url}\fi

\bibitem[Krizhevsky et~al.(2012)Krizhevsky, Sutskever, and
  Hinton]{krizhevsky2012imagenet}
Alex Krizhevsky, Ilya Sutskever, and Geoffrey~E Hinton.
\newblock {Imagenet classification with deep convolutional neural networks}.
\newblock In \emph{Advances in neural information processing systems}, pages
  1097--1105, 2012.

\bibitem[Dodge and Karam(2017)]{Dodge2017}
Samuel Dodge and Lina Karam.
\newblock {A Study and Comparison of Human and Deep Learning Recognition
  Performance Under Visual Distortions}.
\newblock \emph{arXiv preprint arXiv:1705.02498}, 2017.

\bibitem[He et~al.(2015)He, Zhang, Ren, and Sun]{he2015delving}
Kaiming He, Xiangyu Zhang, Shaoqing Ren, and Jian Sun.
\newblock {Delving deep into rectifiers: Surpassing human-level performance on
  imagenet classification}.
\newblock In \emph{Proceedings of the IEEE international conference on computer
  vision}, pages 1026--1034, 2015.

\bibitem[Rajpurkar et~al.(2017)Rajpurkar, Irvin, Zhu, Yang, Mehta, Duan, Ding,
  Bagul, Langlotz, Shpanskaya, and Others]{rajpurkar2017chexnet}
Pranav Rajpurkar, Jeremy Irvin, Kaylie Zhu, Brandon Yang, Hershel Mehta, Tony
  Duan, Daisy Ding, Aarti Bagul, Curtis Langlotz, Katie Shpanskaya, and Others.
\newblock {CheXNet: Radiologist-Level Pneumonia Detection on Chest X-Rays with
  Deep Learning}.
\newblock \emph{arXiv preprint arXiv:1711.05225}, 2017.

\bibitem[Karpathy and Fei-Fei(2015)]{Karpathy2015}
Andrej Karpathy and Li~Fei-Fei.
\newblock {Deep visual-semantic alignments for generating image descriptions}.
\newblock In \emph{Proceedings of the IEEE Conference on Computer Vision and
  Pattern Recognition}, pages 3128--3137, 2015.

\bibitem[Lin et~al.(2017)Lin, Goyal, Girshick, He, and
  Doll{\'{a}}r]{lin2017focal}
Tsung-Yi Lin, Priya Goyal, Ross Girshick, Kaiming He, and Piotr Doll{\'{a}}r.
\newblock {Focal loss for dense object detection}.
\newblock \emph{arXiv preprint arXiv:1708.02002}, 2017.

\bibitem[Ren et~al.(2015)Ren, He, Girshick, and Sun]{Ren2015}
Shaoqing Ren, Kaiming He, Ross Girshick, and Jian Sun.
\newblock {Faster R-CNN: Towards real-time object detection with region
  proposal networks}.
\newblock In \emph{Advances in neural information processing systems}, pages
  91--99, 2015.

\bibitem[Ghate et~al.(2005)Ghate, Soo, Baker, Walsh, Gimenez, and
  Rosen]{Ghate2005}
Sujata~V Ghate, Mary~Scott Soo, Jay~A Baker, Ruth Walsh, Edgardo~I Gimenez, and
  Eric~L Rosen.
\newblock {Comparison of recall and cancer detection rates for immediate versus
  batch interpretation of screening mammograms}.
\newblock \emph{Radiology}, 235\penalty0 (1):\penalty0 31--35, 2005.
\newblock ISSN 0033-8419.

\bibitem[Elsayes et~al.(2017)Elsayes, Hooker, Agrons, Kielar, Tang, Fowler,
  Chernyak, Bashir, Kono, and Do]{Elsayes2017}
Khaled~M Elsayes, Jonathan~C Hooker, Michelle~M Agrons, Ania~Z Kielar, An~Tang,
  Kathryn~J Fowler, Victoria Chernyak, Mustafa~R Bashir, Yuko Kono, and
  Richard~K Do.
\newblock {2017 version of LI-RADS for CT and MR imaging: An update}.
\newblock \emph{RadioGraphics}, 37\penalty0 (7):\penalty0 1994--2017, 2017.
\newblock ISSN 0271-5333.

\bibitem[Mazurowski(2015)]{mazurowski2015radiogenomics}
Maciej~A Mazurowski.
\newblock {Radiogenomics: what it is and why it is important}.
\newblock \emph{Journal of the American College of Radiology}, 12\penalty0
  (8):\penalty0 862--866, 2015.

\bibitem[Mazurowski et~al.(2014)Mazurowski, Zhang, Grimm, Yoon, and
  Silber]{Mazurowski2014}
M~A Mazurowski, J~Zhang, L~J Grimm, S~C Yoon, and J~I Silber.
\newblock {Radiogenomic analysis of breast cancer: Luminal B molecular subtype
  is associated with enhancement dynamics at MR imaging}.
\newblock \emph{Radiology}, 273\penalty0 (2):\penalty0 365--372, 2014.
\newblock \doi{10.1148/radiol.14132641}.

\bibitem[Gutman et~al.(2013)Gutman, Cooper, Hwang, Holder, Gao, Aurora, {Dunn
  Jr}, Scarpace, Mikkelsen, and Jain]{Gutman2013}
David~A Gutman, L~A Cooper, Scott~N Hwang, Chad~A Holder, JingJing Gao, Tarun~D
  Aurora, William~D {Dunn Jr}, Lisa Scarpace, Tom Mikkelsen, and Rajan Jain.
\newblock {MR imaging predictors of molecular profile and survival:
  multi-institutional study of the TCGA glioblastoma data set}.
\newblock \emph{Radiology}, 267\penalty0 (2):\penalty0 560--569, 2013.
\newblock ISSN 0033-8419.

\bibitem[Mazurowski et~al.(2017)Mazurowski, Clark, Czarnek, Shamsesfandabadi,
  Peters, and Saha]{mazurowski2017radiogenomics}
Maciej~A Mazurowski, Kal Clark, Nicholas~M Czarnek, Parisa Shamsesfandabadi,
  Katherine~B Peters, and Ashirbani Saha.
\newblock {Radiogenomics of lower-grade glioma: algorithmically-assessed tumor
  shape is associated with tumor genomic subtypes and patient outcomes in a
  multi-institutional study with The Cancer Genome Atlas data}.
\newblock \emph{Journal of Neuro-Oncology}, pages 1--9, 2017.

\bibitem[Karlo et~al.(2014)Karlo, {Di Paolo}, Chaim, Hakimi, Ostrovnaya, Russo,
  Hricak, Motzer, Hsieh, and Akin]{Karlo2014}
Christoph~A Karlo, Pier~Luigi {Di Paolo}, Joshua Chaim, A~Ari Hakimi, Irina
  Ostrovnaya, Paul Russo, Hedvig Hricak, Robert Motzer, James~J Hsieh, and Oguz
  Akin.
\newblock {Radiogenomics of clear cell renal cell carcinoma: associations
  between CT imaging features and mutations}.
\newblock \emph{Radiology}, 270\penalty0 (2):\penalty0 464--471, 2014.
\newblock ISSN 0033-8419.

\bibitem[Mazurowski et~al.(2013)Mazurowski, Desjardins, and
  Malof]{Mazurowski2013}
Maciej~A Mazurowski, Annick Desjardins, and Jordan~Milton Malof.
\newblock {Imaging descriptors improve the predictive power of survival models
  for glioblastoma patients.}
\newblock \emph{Neuro-oncology}, 15\penalty0 (10):\penalty0 1389--94, oct 2013.
\newblock ISSN 1523-5866.
\newblock \doi{10.1093/neuonc/nos335}.

\bibitem[Mazurowski et~al.(2015)Mazurowski, Grimm, Zhang, Marcom, Yoon, Kim,
  Ghate, and Johnson]{Mazurowski2015Recurrence}
Maciej~A Mazurowski, Lars~J Grimm, Jing Zhang, P~Kelly Marcom, Sora~C Yoon,
  Connie Kim, Sujata~V Ghate, and Karen~S Johnson.
\newblock {Recurrence-free survival in breast cancer is associated with MRI
  tumor enhancement dynamics quantified using computer algorithms.}
\newblock \emph{European journal of radiology}, 84\penalty0 (11):\penalty0
  2117--2122, nov 2015.
\newblock \doi{10.1016/j.ejrad.2015.07.012}.

\bibitem[Erickson et~al.(2017)Erickson, Korfiatis, Akkus, and
  Kline]{Erickson2017}
Bradley~J Erickson, Panagiotis Korfiatis, Zeynettin Akkus, and Timothy~L Kline.
\newblock {Machine Learning for Medical Imaging}.
\newblock \emph{RadioGraphics}, 37\penalty0 (2):\penalty0 505--515, 2017.
\newblock ISSN 0271-5333.

\bibitem[Ruder(2016)]{ruder2016overview}
Sebastian Ruder.
\newblock An overview of gradient descent optimization algorithms.
\newblock \emph{arXiv preprint arXiv:1609.04747}, 2016.

\bibitem[Yosinski et~al.(2014)Yosinski, Clune, Bengio, and
  Lipson]{yosinski2014transferable}
Jason Yosinski, Jeff Clune, Yoshua Bengio, and Hod Lipson.
\newblock {How transferable are features in deep neural networks?}
\newblock In \emph{Advances in neural information processing systems}, pages
  3320--3328, 2014.

\bibitem[{Sharif Razavian} et~al.(2014){Sharif Razavian}, Azizpour, Sullivan,
  and Carlsson]{sharif2014cnn}
Ali {Sharif Razavian}, Hossein Azizpour, Josephine Sullivan, and Stefan
  Carlsson.
\newblock {CNN features off-the-shelf: an astounding baseline for recognition}.
\newblock In \emph{Proceedings of the IEEE conference on computer vision and
  pattern recognition workshops}, pages 806--813, 2014.

\bibitem[Ahmed et~al.(2017)Ahmed, Hall, Goldgof, Liu, and
  Gatenby]{ahmed2017fine}
Kaoutar~B Ahmed, Lawrence~O Hall, Dmitry~B Goldgof, Renhao Liu, and Robert~A
  Gatenby.
\newblock {Fine-tuning convolutional deep features for MRI based brain tumor
  classification}.
\newblock In \emph{Medical Imaging 2017: Computer-Aided Diagnosis}, volume
  10134, page 101342E. International Society for Optics and Photonics, 2017.

\bibitem[Esteva et~al.(2017)Esteva, Kuprel, Novoa, Ko, and Swetter]{Esteva}
A~Esteva, B~Kuprel, RA~Novoa, J~Ko, and SM~Swetter.
\newblock {Dermatologist-level classification of skin cancer with deep neural
  networks}.
\newblock \emph{Nature}, 2017.

\bibitem[Tajbakhsh et~al.(2016)Tajbakhsh, Shin, Gurudu, Hurst, Kendall, Gotway,
  and Liang]{tajbakhsh2016convolutional}
Nima Tajbakhsh, Jae~Y Shin, Suryakanth~R Gurudu, R~Todd Hurst, Christopher~B
  Kendall, Michael~B Gotway, and Jianming Liang.
\newblock {Convolutional neural networks for medical image analysis: Full
  training or fine tuning?}
\newblock \emph{IEEE transactions on medical imaging}, 35\penalty0
  (5):\penalty0 1299--1312, 2016.

\bibitem[Antropova et~al.(2017)Antropova, Huynh, and Giger]{Antropova2017}
Natalia Antropova, Benjamin~Q Huynh, and Maryellen~L Giger.
\newblock A deep feature fusion methodology for breast cancer diagnosis
  demonstrated on three imaging modality datasets.
\newblock \emph{Medical physics}, 2017.

\bibitem[Paul et~al.(2016)Paul, Hawkins, and Balagurunathan]{Paul2016}
R~Paul, SH~Hawkins, and Y~Balagurunathan.
\newblock {Deep Feature Transfer Learning in Combination with Traditional
  Features Predicts Survival Among Patients with Lung Adenocarcinoma}.
\newblock \emph{Tomography}, 2\penalty0 (4):\penalty0 388--395, 2016.
\newblock ISSN 2379-1381.
\newblock \doi{10.18383/j.tom.2016.00211}.

\bibitem[Wang et~al.(2017)Wang, Yang, Weinreb, Han, Li, Kong, Yan, Ke, Luo,
  Liu, and Wang]{Wang2017a}
Xinggang Wang, Wei Yang, Jeffrey Weinreb, Juan Han, Qiubai Li, Xiangchuang
  Kong, Yongluan Yan, Zan Ke, Bo~Luo, Tao Liu, and Liang Wang.
\newblock {Searching for prostate cancer by fully automated magnetic resonance
  imaging classification: Deep learning versus non-deep learning}.
\newblock \emph{Scientific Reports}, 7\penalty0 (1):\penalty0 15415, 2017.
\newblock ISSN 20452322.
\newblock \doi{10.1038/s41598-017-15720-y}.

\bibitem[Lakhani and Sundaram(2017)]{Lakhani2017}
Paras Lakhani and Baskaran Sundaram.
\newblock {Deep Learning at Chest Radiography: Automated Classification of
  Pulmonary Tuberculosis by Using Convolutional Neural Networks}.
\newblock \emph{Radiology}, 284\penalty0 (2):\penalty0 574--582, aug 2017.
\newblock ISSN 0033-8419.
\newblock \doi{10.1148/radiol.2017162326}.

\bibitem[Xue et~al.(2017)Xue, Zhang, Deng, Chen, and Jiang]{Xue2017}
Yanping Xue, Rongguo Zhang, Yufeng Deng, Kuan Chen, and Tao Jiang.
\newblock {A preliminary examination of the diagnostic value of deep learning
  in hip osteoarthritis}.
\newblock \emph{PLOS ONE}, 12\penalty0 (6):\penalty0 e0178992, jun 2017.
\newblock ISSN 1932-6203.
\newblock \doi{10.1371/journal.pone.0178992}.

\bibitem[Chi et~al.(2017)Chi, Walia, Babyn, Wang, Groot, and Eramian]{Chi2017}
Jianning Chi, Ekta Walia, Paul Babyn, Jimmy Wang, Gary Groot, and Mark Eramian.
\newblock {Thyroid Nodule Classification in Ultrasound Images by Fine-Tuning
  Deep Convolutional Neural Network}.
\newblock \emph{Journal of Digital Imaging}, 30\penalty0 (4):\penalty0
  477--486, aug 2017.
\newblock ISSN 1618727X.
\newblock \doi{10.1007/s10278-017-9997-y}.

\bibitem[Kumar et~al.(2017)Kumar, Kim, Lyndon, Fulham, and Feng]{Kumar2017}
A.~Kumar, J.~Kim, D.~Lyndon, M.~Fulham, and D.~Feng.
\newblock {An Ensemble of Fine-Tuned Convolutional Neural Networks for Medical
  Image Classification}.
\newblock \emph{IEEE Journal of Biomedical and Health Informatics}, 21\penalty0
  (1):\penalty0 31--40, 2017.
\newblock ISSN 2168-2194.
\newblock \doi{10.1109/JBHI.2016.2635663}.

\bibitem[Zhu et~al.(2017{\natexlab{a}})Zhu, Albadawy, Saha, Zhang, Harowicz,
  and Mazurowski]{Zhu2017a}
Zhe Zhu, Ehab Albadawy, Ashirbani Saha, Jun Zhang, Michael~R. Harowicz, and
  Maciej~A. Mazurowski.
\newblock {Deep Learning for identifying radiogenomic associations in breast
  cancer}.
\newblock \emph{arXiv preprint arXiv:1711.11097}, 2017{\natexlab{a}}.

\bibitem[Zhu et~al.(2017{\natexlab{b}})Zhu, Harowicz, Zhang, Saha, Grimm,
  Hwang, and Mazurowski]{Zhu2017}
Zhe Zhu, Michael~R. Harowicz, Jun Zhang, Ashirbani Saha, Lars~J. Grimm,
  E.~Shelley Hwang, and Maciej~A. Mazurowski.
\newblock {Deep learning analysis of breast MRIs for prediction of occult
  invasive disease in ductal carcinoma in situ}.
\newblock \emph{arXiv:1711.10577}, 2017{\natexlab{b}}.

\bibitem[Simonyan and Zisserman(2014)]{simonyan2014very}
Karen Simonyan and Andrew Zisserman.
\newblock {Very deep convolutional networks for large-scale image recognition}.
\newblock \emph{arXiv preprint arXiv:1409.1556}, 2014.

\bibitem[Li et~al.(2017{\natexlab{a}})Li, Wang, Yu, Guo, and Cao]{li2017deep}
Zeju Li, Yuanyuan Wang, Jinhua Yu, Yi~Guo, and Wei Cao.
\newblock {Deep Learning based Radiomics (DLR) and its usage in noninvasive
  IDH1 prediction for low grade glioma}.
\newblock \emph{Scientific Reports}, 7\penalty0 (1), 2017{\natexlab{a}}.
\newblock ISSN 20452322.
\newblock \doi{10.1038/s41598-017-05848-2}.

\bibitem[Suk et~al.(2017)Suk, Lee, and Shen]{Suk2017}
Heung-Il Suk, Seong-Whan Lee, and Dinggang Shen.
\newblock {Deep ensemble learning of sparse regression models for brain disease
  diagnosis}.
\newblock \emph{Medical Image Analysis}, 37:\penalty0 101--113, 2017.
\newblock ISSN 13618415.
\newblock \doi{10.1016/j.media.2017.01.008}.

\bibitem[Khawaldeh et~al.()Khawaldeh, Pervaiz, Rafiq, and Sciences]{Khawaldeh}
S~Khawaldeh, U~Pervaiz, A~Rafiq, and RS~Alkhawaldeh Sciences.
\newblock {Noninvasive Grading of Glioma Tumor Using Magnetic Resonance Imaging
  with Convolutional Neural Networks}.
\newblock \emph{mdpi.com}.

\bibitem[Gonz{\'{a}}lez et~al.(2017)Gonz{\'{a}}lez, Ash, {Vegas
  Sanchez-Ferrero}, {Onieva Onieva}, Rahaghi, Ross, D{\'{i}}az, {San Jos{\'{e}}
  Est{\'{e}}par}, Washko, {COPDGene and ECLIPSE investigators}, Zone, and
  Paper]{Gonzalez2017}
Germ{\'{a}}n Gonz{\'{a}}lez, Samuel~Y Ash, Gonzalo {Vegas Sanchez-Ferrero},
  Jorge {Onieva Onieva}, Farbod~N Rahaghi, James~C Ross, Alejandro D{\'{i}}az,
  Ra{\'{u}}l {San Jos{\'{e}} Est{\'{e}}par}, George~R Washko, {COPDGene and
  ECLIPSE investigators}, Vadose Zone, and J~Accepted Paper.
\newblock {Disease Staging and Prognosis in Smokers using Deep Learning in
  Chest Computed Tomography}.
\newblock \emph{American Journal of Respiratory and Critical Care Medicine},
  pages 1--49, sep 2017.
\newblock ISSN 1073-449X.
\newblock \doi{10.2136/vzj2016.06.0055}.

\bibitem[Canziani et~al.(2016)Canziani, Paszke, and
  Culurciello]{canziani2016analysis}
Alfredo Canziani, Adam Paszke, and Eugenio Culurciello.
\newblock {An analysis of deep neural network models for practical
  applications}.
\newblock \emph{arXiv preprint arXiv:1605.07678}, 2016.

\bibitem[He et~al.(2016)He, Zhang, Ren, and Sun]{he2016deep}
Kaiming He, Xiangyu Zhang, Shaoqing Ren, and Jian Sun.
\newblock {Deep residual learning for image recognition}.
\newblock In \emph{Proceedings of the IEEE Conference on Computer Vision and
  Pattern Recognition}, pages 770--778, 2016.

\bibitem[Szegedy et~al.(2017)Szegedy, Ioffe, Vanhoucke, and
  Alemi]{szegedy2017inception}
Christian Szegedy, Sergey Ioffe, Vincent Vanhoucke, and Alexander~A Alemi.
\newblock {Inception-v4, Inception-ResNet and the Impact of Residual
  Connections on Learning.}
\newblock In \emph{AAAI}, pages 4278--4284, 2017.

\bibitem[Szegedy et~al.(2016)Szegedy, Vanhoucke, Ioffe, Shlens, and
  Wojna]{szegedy2016rethinking}
Christian Szegedy, Vincent Vanhoucke, Sergey Ioffe, Jon Shlens, and Zbigniew
  Wojna.
\newblock {Rethinking the inception architecture for computer vision}.
\newblock In \emph{Proceedings of the IEEE Conference on Computer Vision and
  Pattern Recognition}, pages 2818--2826, 2016.

\bibitem[Korfiatis et~al.(2017)Korfiatis, Kline, Lachance, Parney, Buckner, and
  Erickson]{Korfiatis2017}
Panagiotis Korfiatis, Timothy~L. Kline, Daniel~H. Lachance, Ian~F. Parney,
  Jan~C. Buckner, and Bradley~J. Erickson.
\newblock {Residual Deep Convolutional Neural Network Predicts MGMT Methylation
  Status}.
\newblock \emph{Journal of Digital Imaging}, 30\penalty0 (5):\penalty0
  622--628, oct 2017.
\newblock ISSN 1618727X.
\newblock \doi{10.1007/s10278-017-0009-z}.

\bibitem[Kim and Mackinnon(2017)]{Kim2017}
D~H Kim and T~Mackinnon.
\newblock {Artificial intelligence in fracture detection: transfer learning
  from deep convolutional neural networks}.
\newblock \emph{Elsevier}, 2017.
\newblock ISSN 1365229X.
\newblock \doi{10.1016/j.crad.2017.11.015}.

\bibitem[Hinton and Salakhutdinov(2006)]{hinton2006reducing}
Geoffrey~E Hinton and Ruslan~R Salakhutdinov.
\newblock {Reducing the dimensionality of data with neural networks}.
\newblock \emph{science}, 313\penalty0 (5786):\penalty0 504--507, 2006.

\bibitem[Bengio et~al.(2007)Bengio, Lamblin, Popovici, and
  Larochelle]{bengio2007greedy}
Yoshua Bengio, Pascal Lamblin, Dan Popovici, and Hugo Larochelle.
\newblock {Greedy layer-wise training of deep networks}.
\newblock In \emph{Advances in neural information processing systems}, pages
  153--160, 2007.

\bibitem[Poultney et~al.(2007)Poultney, Chopra, Cun, and
  Others]{poultney2007efficient}
Christopher Poultney, Sumit Chopra, Yann~L Cun, and Others.
\newblock {Efficient learning of sparse representations with an energy-based
  model}.
\newblock In \emph{Advances in neural information processing systems}, pages
  1137--1144, 2007.

\bibitem[Ortiz et~al.(2017)Ortiz, Munilla, Mart{\'{i}}nez-Murcia, G{\'{o}}rriz,
  and Ram{\'{i}}rez]{Ortiz2017}
Andr{\'{e}}s Ortiz, Jorge Munilla, Francisco~J. Mart{\'{i}}nez-Murcia, Juan~M.
  G{\'{o}}rriz, and Javier Ram{\'{i}}rez.
\newblock {Learning longitudinal MRI patterns by SICE and deep learning:
  Assessing the Alzheimer's disease progression}.
\newblock In \emph{Communications in Computer and Information Science}, volume
  723, pages 413--424, 2017.
\newblock ISBN 9783319609638.
\newblock \doi{10.1007/978-3-319-60964-5_36}.

\bibitem[Kumar et~al.(2015)Kumar, Wong, and Clausi]{Kumar2015}
Devinder Kumar, Alexander Wong, and David~A. Clausi.
\newblock {Lung Nodule Classification Using Deep Features in CT Images}.
\newblock In \emph{2015 12th Conference on Computer and Robot Vision}, pages
  133--138, 2015.
\newblock ISBN 978-1-4799-1986-4.
\newblock \doi{10.1109/CRV.2015.25}.

\bibitem[Yoo et~al.(2018)Yoo, Tang, Brosch, Li, Kolind, Vavasour, Rauscher,
  MacKay, Traboulsee, and Tam]{Yoo2018}
Youngjin Yoo, Lisa~Y.W. Tang, Tom Brosch, David~K.B. Li, Shannon Kolind, Irene
  Vavasour, Alexander Rauscher, Alex~L. MacKay, Anthony Traboulsee, and
  Roger~C. Tam.
\newblock {Deep learning of joint myelin and T1w MRI features in
  normal-appearing brain tissue to distinguish between multiple sclerosis
  patients and healthy controls}.
\newblock \emph{NeuroImage: Clinical}, 17:\penalty0 169--178, 2018.
\newblock ISSN 22131582.
\newblock \doi{10.1016/j.nicl.2017.10.015}.

\bibitem[Wang and Summers(2012)]{Wang2012}
Shijun Wang and Ronald~M Summers.
\newblock {Machine learning and radiology}.
\newblock \emph{Medical Image Analysis}, 16\penalty0 (5):\penalty0 933--951,
  2012.
\newblock ISSN 1361-8415.
\newblock \doi{https://doi.org/10.1016/j.media.2012.02.005}.

\bibitem[Kim(2014)]{Kim2014}
Yoon Kim.
\newblock {Convolutional Neural Networks for Sentence Classification}.
\newblock In \emph{Proceedings of the 2014 Conference on Empirical Methods in
  Natural Language Processing (EMNLP), Doha, Qatar, October 25–29, 2014.
  Stroudsburg}, pages 1746--1751. Citeseer, 2014.

\bibitem[Mikolov et~al.(2013)Mikolov, Sutskever, Chen, Corrado, and
  Dean]{Mikolov2013}
Tomas Mikolov, Ilya Sutskever, Kai Chen, Greg~S Corrado, and Jeff Dean.
\newblock {Distributed representations of words and phrases and their
  compositionality}.
\newblock In \emph{Advances in neural information processing systems}, pages
  3111--3119, 2013.

\bibitem[Chen et~al.(2017{\natexlab{a}})Chen, Ball, Yang, Moradzadeh, Chapman,
  Larson, Langlotz, Amrhein, and Lungren]{Chen2017a}
Matthew~C. Chen, Robyn~L. Ball, Lingyao Yang, Nathaniel Moradzadeh, Brian~E.
  Chapman, David~B. Larson, Curtis~P. Langlotz, Timothy~J. Amrhein, and
  Matthew~P. Lungren.
\newblock {Deep Learning to Classify Radiology Free-Text Reports}.
\newblock \emph{Radiology}, page 171115, nov 2017{\natexlab{a}}.
\newblock ISSN 0033-8419.
\newblock \doi{10.1148/radiol.2017171115}.

\bibitem[Shin et~al.(2017)Shin, Chokshi, Lee, and Choi]{Shin2017}
Bonggun Shin, Falgun~H. Chokshi, Timothy Lee, and Jinho~D. Choi.
\newblock {Classification of radiology reports using neural attention models}.
\newblock In \emph{Proceedings of the International Joint Conference on Neural
  Networks}, volume 2017-May, pages 4363--4370, 2017.
\newblock ISBN 9781509061815.
\newblock \doi{10.1109/IJCNN.2017.7966408}.

\bibitem[Karimi et~al.(2017)Karimi, Dai, Hassanzadeh, and Nguyen]{Karimi2017}
Sarvnaz Karimi, Xiang Dai, Hamedh Hassanzadeh, and Anthony Nguyen.
\newblock {Automatic Diagnosis Coding of Radiology Reports: A Comparison of
  Deep Learning and Conventional Classification Methods}.
\newblock \emph{BioNLP}, pages 328--332, 2017.

\bibitem[Akkus et~al.(2017)Akkus, Ali, Sedl{\'{a}}ř, Agrawal, Parney,
  Giannini, and Erickson]{akkus2017predicting}
Zeynettin Akkus, Issa Ali, Jiř{\'{i}} Sedl{\'{a}}ř, Jay~P. Agrawal, Ian~F.
  Parney, Caterina Giannini, and Bradley~J. Erickson.
\newblock {Predicting Deletion of Chromosomal Arms 1p/19q in Low-Grade Gliomas
  from MR Images Using Machine Intelligence}.
\newblock \emph{Journal of Digital Imaging}, 30\penalty0 (4):\penalty0
  469--476, aug 2017.
\newblock ISSN 0897-1889.
\newblock \doi{10.1007/s10278-017-9984-3}.

\bibitem[Havaei et~al.(2015)Havaei, Davy, Warde-Farley, Biard, Courville,
  Bengio, Pal, Jodoin, and Larochelle]{Havaei2015}
Mohammad Havaei, Axel Davy, David Warde-Farley, Antoine Biard, Aaron Courville,
  Yoshua Bengio, Chris Pal, Pierre-Marc Jodoin, and Hugo Larochelle.
\newblock {Brain Tumor Segmentation with Deep Neural Networks}.
\newblock \emph{arXiv preprint arXiv:1505.03540}, 2015.

\bibitem[Hussain et~al.(2017)Hussain, Anwar, and Majid]{hussain2017brain}
Saddam Hussain, Syed~Muhammad Anwar, and Muhammad Majid.
\newblock {Brain tumor segmentation using cascaded deep convolutional neural
  network}.
\newblock In \emph{Engineering in Medicine and Biology Society (EMBC), 2017
  39th Annual International Conference of the IEEE}, pages 1998--2001. IEEE,
  2017.

\bibitem[Milletari et~al.(2017)Milletari, Ahmadi, Kroll, Plate, Rozanski,
  Maiostre, Levin, Dietrich, Ertl-Wagner, B{\"{o}}tzel, and
  Others]{milletari2017hough}
Fausto Milletari, Seyed-Ahmad Ahmadi, Christine Kroll, Annika Plate, Verena
  Rozanski, Juliana Maiostre, Johannes Levin, Olaf Dietrich, Birgit
  Ertl-Wagner, Kai B{\"{o}}tzel, and Others.
\newblock {Hough-CNN: Deep learning for segmentation of deep brain regions in
  MRI and ultrasound}.
\newblock \emph{Computer Vision and Image Understanding}, 2017.

\bibitem[Valverde et~al.(2017)Valverde, Cabezas, Roura,
  Gonz{\'{a}}lez-Vill{\`{a}}, Pareto, Vilanova, Rami{\'{o}}-Torrent{\`{a}},
  Rovira, Oliver, and Llad{\'{o}}]{valverde2017improving}
Sergi Valverde, Mariano Cabezas, Eloy Roura, Sandra Gonz{\'{a}}lez-Vill{\`{a}},
  Deborah Pareto, Joan~C Vilanova, Llu$\backslash$'$\backslash$is
  Rami{\'{o}}-Torrent{\`{a}}, {\`{A}}lex Rovira, Arnau Oliver, and Xavier
  Llad{\'{o}}.
\newblock {Improving automated multiple sclerosis lesion segmentation with a
  cascaded 3D convolutional neural network approach}.
\newblock \emph{NeuroImage}, 155:\penalty0 159--168, 2017.

\bibitem[Wachinger et~al.(2017)Wachinger, Reuter, and
  Klein]{wachinger2017deepnat}
Christian Wachinger, Martin Reuter, and Tassilo Klein.
\newblock {DeepNAT: Deep convolutional neural network for segmenting
  neuroanatomy}.
\newblock \emph{NeuroImage}, 2017.

\bibitem[Trebeschi et~al.(2017)Trebeschi, van Griethuysen, Lambregts, Lahaye,
  Parmer, Bakers, Peters, Beets-Tan, and Aerts]{trebeschi2017deep}
Stefano Trebeschi, Joost J~M van Griethuysen, Doenja M~J Lambregts, Max~J
  Lahaye, Chintan Parmer, Frans C~H Bakers, Nicky H G~M Peters, Regina G~H
  Beets-Tan, and Hugo J W~L Aerts.
\newblock {Deep learning for fully-automated localization and segmentation of
  rectal cancer on multiparametric MR}.
\newblock \emph{Scientific reports}, 7\penalty0 (1):\penalty0 5301, 2017.

\bibitem[Long et~al.(2015)Long, Shelhamer, and Darrell]{long2015fully}
Jonathan Long, Evan Shelhamer, and Trevor Darrell.
\newblock {Fully convolutional networks for semantic segmentation}.
\newblock In \emph{Proceedings of the IEEE Conference on Computer Vision and
  Pattern Recognition}, pages 3431--3440, 2015.

\bibitem[Christ et~al.(2017)Christ, Ettlinger, Gr{\"{u}}n, Elshaera, Lipkova,
  Schlecht, Ahmaddy, Tatavarty, Bickel, Bilic, and Others]{christ2017automatic}
Patrick~Ferdinand Christ, Florian Ettlinger, Felix Gr{\"{u}}n, Mohamed
  Ezzeldin~A Elshaera, Jana Lipkova, Sebastian Schlecht, Freba Ahmaddy, Sunil
  Tatavarty, Marc Bickel, Patrick Bilic, and Others.
\newblock {Automatic Liver and Tumor Segmentation of CT and MRI Volumes using
  Cascaded Fully Convolutional Neural Networks}.
\newblock \emph{arXiv preprint arXiv:1702.05970}, 2017.

\bibitem[Ronneberger et~al.(2015)Ronneberger, Fischer, and
  Brox]{ronneberger2015u}
Olaf Ronneberger, Philipp Fischer, and Thomas Brox.
\newblock {U-net: Convolutional networks for biomedical image segmentation}.
\newblock In \emph{International Conference on Medical Image Computing and
  Computer-Assisted Intervention}, pages 234--241. Springer, 2015.

\bibitem[Salehi et~al.(2017)Salehi, Hashemi, Velasco-Annis, Ouaalam, Estroff,
  Erdogmus, Warfield, and Gholipour]{salehi2017real}
Seyed Sadegh~Mohseni Salehi, Seyed~Raein Hashemi, Clemente Velasco-Annis,
  Abdelhakim Ouaalam, Judy~A Estroff, Deniz Erdogmus, Simon~K Warfield, and Ali
  Gholipour.
\newblock {Real-Time Automatic Fetal Brain Extraction in Fetal MRI by Deep
  Learning}.
\newblock \emph{arXiv preprint arXiv:1710.09338}, 2017.

\bibitem[J{\'{e}}gou et~al.(2017)J{\'{e}}gou, Drozdzal, Vazquez, Romero, and
  Bengio]{jegou2017one}
Simon J{\'{e}}gou, Michal Drozdzal, David Vazquez, Adriana Romero, and Yoshua
  Bengio.
\newblock {The one hundred layers tiramisu: Fully convolutional densenets for
  semantic segmentation}.
\newblock In \emph{Computer Vision and Pattern Recognition Workshops (CVPRW),
  2017 IEEE Conference on}, pages 1175--1183. IEEE, 2017.

\bibitem[Chen et~al.()Chen, Wu, DSouza, Abidin, Xu, and
  Wism{\"{u}}ller]{chenmri}
Lele Chen, Yue Wu, Adora~M DSouza, Anas~Z Abidin, Chenliang Xu, and Axel
  Wism{\"{u}}ller.
\newblock {MRI tumor segmentation with densely connected 3D CNN}.

\bibitem[Li et~al.(2017{\natexlab{b}})Li, Chen, Qi, Dou, Fu, and Heng]{li2017h}
Xiaomeng Li, Hao Chen, Xiaojuan Qi, Qi~Dou, Chi-Wing Fu, and Pheng~Ann Heng.
\newblock {H-DenseUNet: Hybrid densely connected UNet for liver and liver tumor
  segmentation from CT volumes}.
\newblock \emph{arXiv preprint arXiv:1709.07330}, 2017{\natexlab{b}}.

\bibitem[Clark et~al.(2017)Clark, Wong, Haider, and Khalvati]{clark2017fully}
Tyler Clark, Alexander Wong, Masoom~A Haider, and Farzad Khalvati.
\newblock {Fully deep convolutional neural networks for segmentation of the
  prostate gland in diffusion-weighted MR images}.
\newblock In \emph{International Conference Image Analysis and Recognition},
  pages 97--104. Springer, 2017.

\bibitem[McKinley et~al.(2016)McKinley, Wepfer, Gundersen, Wagner, Chan, Wiest,
  and Reyes]{mckinley2016nabla}
Richard McKinley, Rik Wepfer, Tom Gundersen, Franca Wagner, Andrew Chan, Roland
  Wiest, and Mauricio Reyes.
\newblock {Nabla-net: A Deep Dag-Like Convolutional Architecture for Biomedical
  Image Segmentation}.
\newblock In \emph{International Workshop on Brainlesion: Glioma, Multiple
  Sclerosis, Stroke and Traumatic Brain Injuries}, pages 119--128. Springer,
  2016.

\bibitem[Zhang et~al.(2017{\natexlab{a}})Zhang, Cui, Niu, Geng, and
  Qiao]{zhang2017image}
Qiao Zhang, Zhipeng Cui, Xiaoguang Niu, Shijie Geng, and Yu~Qiao.
\newblock {Image Segmentation with Pyramid Dilated Convolution Based on ResNet
  and U-Net}.
\newblock In \emph{International Conference on Neural Information Processing},
  pages 364--372. Springer, 2017{\natexlab{a}}.

\bibitem[Mehta and Sivaswamy(2017)]{mehta2017m}
Raghav Mehta and Jayanthi Sivaswamy.
\newblock {M-net: A Convolutional Neural Network for deep brain structure
  segmentation}.
\newblock In \emph{Biomedical Imaging (ISBI 2017), 2017 IEEE 14th International
  Symposium on}, pages 437--440. IEEE, 2017.

\bibitem[Sudre et~al.(2017)Sudre, Li, Vercauteren, Ourselin, and
  Cardoso]{sudre2017generalised}
Carole~H Sudre, Wenqi Li, Tom Vercauteren, Sebastien Ourselin, and M~Jorge
  Cardoso.
\newblock {Generalised Dice overlap as a deep learning loss function for highly
  unbalanced segmentations}.
\newblock In \emph{Deep Learning in Medical Image Analysis and Multimodal
  Learning for Clinical Decision Support}, pages 240--248. Springer, 2017.

\bibitem[Milletari et~al.(2016)Milletari, Navab, and Ahmadi]{milletari2016v}
Fausto Milletari, Nassir Navab, and Seyed-Ahmad Ahmadi.
\newblock {V-net: Fully convolutional neural networks for volumetric medical
  image segmentation}.
\newblock In \emph{3D Vision (3DV), 2016 Fourth International Conference on},
  pages 565--571. IEEE, 2016.

\bibitem[{\c{C}}i{\c{c}}ek et~al.(2016){\c{C}}i{\c{c}}ek, Abdulkadir, Lienkamp,
  Brox, and Ronneberger]{cciccek20163d}
{\"{O}}zg{\"{u}}n {\c{C}}i{\c{c}}ek, Ahmed Abdulkadir, Soeren~S Lienkamp,
  Thomas Brox, and Olaf Ronneberger.
\newblock {3D U-Net: learning dense volumetric segmentation from sparse
  annotation}.
\newblock In \emph{International Conference on Medical Image Computing and
  Computer-Assisted Intervention}, pages 424--432. Springer, 2016.

\bibitem[Deniz et~al.(2017)Deniz, Hallyburton, Welbeck, Honig, Cho, and
  Chang]{deniz2017segmentation}
Cem~M Deniz, Spencer Hallyburton, Arakua Welbeck, Stephen Honig, Kyunghyun Cho,
  and Gregory Chang.
\newblock {Segmentation of the Proximal Femur from MR Images using Deep
  Convolutional Neural Networks}.
\newblock \emph{arXiv preprint arXiv:1704.06176}, 2017.

\bibitem[Shen and Anderson()]{shenmultimodal}
Liyue Shen and Timothy Anderson.
\newblock {Multimodal Brain MRI Tumor Segmentation via Convolutional Neural
  Networks}.

\bibitem[Yang et~al.(2017{\natexlab{a}})Yang, Yu, Wu, Wang, Ni, Qin, and
  Heng]{yang2017fine}
Xin Yang, Lequan Yu, Lingyun Wu, Yi~Wang, Dong Ni, Jing Qin, and Pheng-Ann
  Heng.
\newblock {Fine-Grained Recurrent Neural Networks for Automatic Prostate
  Segmentation in Ultrasound Images.}
\newblock In \emph{AAAI}, pages 1633--1639, 2017{\natexlab{a}}.

\bibitem[Poudel et~al.(2016)Poudel, Lamata, and Montana]{poudel2016recurrent}
Rudra P~K Poudel, Pablo Lamata, and Giovanni Montana.
\newblock {Recurrent fully convolutional neural networks for multi-slice mri
  cardiac segmentation}.
\newblock In \emph{International Workshop on Reconstruction and Analysis of
  Moving Body Organs}, pages 83--94. Springer, 2016.

\bibitem[Cai et~al.(2017)Cai, Lu, Xie, Xing, and Yang]{cai2017improving}
Jinzheng Cai, Le~Lu, Yuanpu Xie, Fuyong Xing, and Lin Yang.
\newblock {Improving deep pancreas segmentation in ct and mri images via
  recurrent neural contextual learning and direct loss function}.
\newblock \emph{arXiv preprint arXiv:1707.04912}, 2017.

\bibitem[Hochreiter and Schmidhuber(1997)]{hochreiter1997long}
Sepp Hochreiter and J{\"{u}}rgen Schmidhuber.
\newblock {Long short-term memory}.
\newblock \emph{Neural computation}, 9\penalty0 (8):\penalty0 1735--1780, 1997.

\bibitem[Jaccard(1912)]{jaccard1912distribution}
Paul Jaccard.
\newblock {The distribution of the flora in the alpine zone.}
\newblock \emph{New phytologist}, 11\penalty0 (2):\penalty0 37--50, 1912.

\bibitem[Al-Kofahi et~al.(2010)Al-Kofahi, Lassoued, Lee, and
  Roysam]{al2010improved}
Yousef Al-Kofahi, Wiem Lassoued, William Lee, and Badrinath Roysam.
\newblock {Improved automatic detection and segmentation of cell nuclei in
  histopathology images}.
\newblock \emph{IEEE Transactions on Biomedical Engineering}, 57\penalty0
  (4):\penalty0 841--852, 2010.

\bibitem[Oliver et~al.(2010)Oliver, Freixenet, Marti, P{\'{e}}rez, Pont,
  Denton, and Zwiggelaar]{oliver2010review}
Arnau Oliver, Jordi Freixenet, Joan Marti, Elsa P{\'{e}}rez, Josep Pont, Erika
  R~E Denton, and Reyer Zwiggelaar.
\newblock {A review of automatic mass detection and segmentation in
  mammographic images}.
\newblock \emph{Medical image analysis}, 14\penalty0 (2):\penalty0 87--110,
  2010.

\bibitem[Rey et~al.(2002)Rey, Subsol, Delingette, and Ayache]{rey2002automatic}
David Rey, G{\'{e}}rard Subsol, Herv{\'{e}} Delingette, and Nicholas Ayache.
\newblock {Automatic detection and segmentation of evolving processes in 3D
  medical images: Application to multiple sclerosis}.
\newblock \emph{Medical image analysis}, 6\penalty0 (2):\penalty0 163--179,
  2002.

\bibitem[Roth et~al.(2014)Roth, Lu, Seff, Cherry, Hoffman, Wang, Liu, Turkbey,
  and Summers]{roth2014new}
Holger~R Roth, Le~Lu, Ari Seff, Kevin~M Cherry, Joanne Hoffman, Shijun Wang,
  Jiamin Liu, Evrim Turkbey, and Ronald~M Summers.
\newblock {A new 2.5 D representation for lymph node detection using random
  sets of deep convolutional neural network observations}.
\newblock In \emph{International Conference on Medical Image Computing and
  Computer-Assisted Intervention}, pages 520--527. Springer, 2014.

\bibitem[Erhan et~al.(2014)Erhan, Szegedy, Toshev, and
  Anguelov]{erhan2014scalable}
Dumitru Erhan, Christian Szegedy, Alexander Toshev, and Dragomir Anguelov.
\newblock {Scalable object detection using deep neural networks}.
\newblock In \emph{Proceedings of the IEEE Conference on Computer Vision and
  Pattern Recognition}, pages 2147--2154, 2014.

\bibitem[Szegedy et~al.(2014)Szegedy, Reed, Erhan, Anguelov, and
  Ioffe]{szegedy2014scalable}
Christian Szegedy, Scott Reed, Dumitru Erhan, Dragomir Anguelov, and Sergey
  Ioffe.
\newblock {Scalable, high-quality object detection}.
\newblock \emph{arXiv preprint arXiv:1412.1441}, 2014.

\bibitem[Shin et~al.(2016)Shin, Roth, Gao, Lu, Xu, Nogues, Yao, Mollura, and
  Summers]{shin2016deep}
Hoo-Chang Shin, Holger~R Roth, Mingchen Gao, Le~Lu, Ziyue Xu, Isabella Nogues,
  Jianhua Yao, Daniel Mollura, and Ronald~M Summers.
\newblock {Deep convolutional neural networks for computer-aided detection: CNN
  architectures, dataset characteristics and transfer learning}.
\newblock \emph{IEEE transactions on medical imaging}, 35\penalty0
  (5):\penalty0 1285--1298, 2016.

\bibitem[Samala et~al.(2016)Samala, Chan, Hadjiiski, Helvie, Wei, and
  Cha]{samala2016mass}
Ravi~K Samala, Heang-Ping Chan, Lubomir Hadjiiski, Mark~A Helvie, Jun Wei, and
  Kenny Cha.
\newblock {Mass detection in digital breast tomosynthesis: Deep convolutional
  neural network with transfer learning from mammography}.
\newblock \emph{Medical physics}, 43\penalty0 (12):\penalty0 6654--6666, 2016.

\bibitem[Szegedy et~al.(2015)Szegedy, Liu, Jia, Sermanet, Reed, Anguelov,
  Erhan, Vanhoucke, and Rabinovich]{szegedy2015going}
Christian Szegedy, Wei Liu, Yangqing Jia, Pierre Sermanet, Scott Reed, Dragomir
  Anguelov, Dumitru Erhan, Vincent Vanhoucke, and Andrew Rabinovich.
\newblock {Going deeper with convolutions}.
\newblock In \emph{Proceedings of the IEEE conference on computer vision and
  pattern recognition}, pages 1--9, 2015.

\bibitem[Sa et~al.(2017)Sa, Owens, Wiegand, Studin, Capoferri, Barooha, Greaux,
  Rattray, Hutton, Cintineo, and Others]{sa2017intervertebral}
Ruhan Sa, William Owens, Raymond Wiegand, Mark Studin, Donald Capoferri,
  Kenneth Barooha, Alexander Greaux, Robert Rattray, Adam Hutton, John
  Cintineo, and Others.
\newblock {Intervertebral disc detection in X-ray images using faster R-CNN}.
\newblock In \emph{Engineering in Medicine and Biology Society (EMBC), 2017
  39th Annual International Conference of the IEEE}, pages 564--567. IEEE,
  2017.

\bibitem[Liu et~al.(2017)Liu, Wang, Lu, Wei, Kim, Turkbey, Sahiner, Petrick,
  and Summers]{liu2017detection}
Jiamin Liu, David Wang, Le~Lu, Zhuoshi Wei, Lauren Kim, Evrim~B Turkbey,
  Berkman Sahiner, Nicholas Petrick, and Ronald~M Summers.
\newblock detection and diagnosis of colitis on computed tomography using deep
  convolutional neural networks.
\newblock \emph{Medical Physics}, 2017.

\bibitem[Ben-Ari et~al.(2017)Ben-Ari, Akselrod-Ballin, Karlinsky, and
  Hashoul]{ben2017domain}
Rami Ben-Ari, Ayelet Akselrod-Ballin, Leonid Karlinsky, and Sharbell Hashoul.
\newblock {Domain specific convolutional neural nets for detection of
  architectural distortion in mammograms}.
\newblock In \emph{Biomedical Imaging (ISBI 2017), 2017 IEEE 14th International
  Symposium on}, pages 552--556. IEEE, 2017.

\bibitem[Redmon et~al.(2016)Redmon, Divvala, Girshick, and
  Farhadi]{redmon2016you}
Joseph Redmon, Santosh Divvala, Ross Girshick, and Ali Farhadi.
\newblock {You only look once: Unified, real-time object detection}.
\newblock In \emph{Proceedings of the IEEE Conference on Computer Vision and
  Pattern Recognition}, pages 779--788, 2016.

\bibitem[Liu et~al.(2016)Liu, Anguelov, Erhan, Szegedy, Reed, Fu, and
  Berg]{liu2016ssd}
Wei Liu, Dragomir Anguelov, Dumitru Erhan, Christian Szegedy, Scott Reed,
  Cheng-Yang Fu, and Alexander~C Berg.
\newblock {Ssd: Single shot multibox detector}.
\newblock In \emph{European conference on computer vision}, pages 21--37.
  Springer, 2016.

\bibitem[Platania et~al.(2017)Platania, Shams, Yang, Zhang, Lee, and
  Park]{platania2017automated}
Richard Platania, Shayan Shams, Seungwon Yang, Jian Zhang, Kisung Lee, and
  Seung-Jong Park.
\newblock {Automated Breast Cancer Diagnosis Using Deep Learning and Region of
  Interest Detection (BC-DROID)}.
\newblock In \emph{Proceedings of the 8th ACM International Conference on
  Bioinformatics, Computational Biology, and Health Informatics}, pages
  536--543. ACM, 2017.

\bibitem[Cao et~al.(2017)Cao, Duan, Yang, Yue, Chen, Fu, and Xu]{cao2017breast}
Zhantao Cao, Lixin Duan, Guowu Yang, Ting Yue, Qin Chen, Huazhu Fu, and Yanwu
  Xu.
\newblock {Breast Tumor Detection in Ultrasound Images Using Deep Learning}.
\newblock In \emph{International Workshop on Patch-based Techniques in Medical
  Imaging}, pages 121--128. Springer, 2017.

\bibitem[Li et~al.(2017{\natexlab{c}})Li, Liu, Qiu, Guo, Zhao, Li, and
  He]{li2017detection}
Ning Li, Haopeng Liu, Bin Qiu, Wei Guo, Shijun Zhao, Kungang Li, and Jie He.
\newblock {Detection and Attention: Diagnosing Pulmonary Lung Cancer from CT by
  Imitating Physicians}.
\newblock \emph{arXiv preprint arXiv:1712.05114}, 2017{\natexlab{c}}.

\bibitem[de~Vos et~al.(2016)de~Vos, Wolterink, de~Jong, Viergever, and
  Isgum]{de20162d}
Bob~D de~Vos, Jelmer~M Wolterink, Pim~A de~Jong, Max~A Viergever, and Ivana
  Isgum.
\newblock {2D image classification for 3D anatomy localization: employing deep
  convolutional neural networks.}
\newblock In \emph{Medical Imaging: Image Processing}, page 97841Y, 2016.

\bibitem[Prasoon et~al.(2013)Prasoon, Petersen, Igel, Lauze, Dam, and
  Nielsen]{prasoon2013deep}
Adhish Prasoon, Kersten Petersen, Christian Igel, Fran{\c{c}}ois Lauze, Erik
  Dam, and Mads Nielsen.
\newblock {Deep feature learning for knee cartilage segmentation using a
  triplanar convolutional neural network}.
\newblock In \emph{International conference on medical image computing and
  computer-assisted intervention}, pages 246--253. Springer, 2013.

\bibitem[Roth et~al.(2016{\natexlab{a}})Roth, Wang, Yao, Lu, Burns, and
  Summers]{roth2016deep}
Holger~R Roth, Yinong Wang, Jianhua Yao, Le~Lu, Joseph~E Burns, and Ronald~M
  Summers.
\newblock {Deep convolutional networks for automated detection of
  posterior-element fractures on spine CT}.
\newblock \emph{arXiv preprint arXiv:1602.00020}, 2016{\natexlab{a}}.

\bibitem[Roth et~al.(2016{\natexlab{b}})Roth, Lu, Liu, Yao, Seff, Cherry, Kim,
  and Summers]{roth2016improving}
Holger~R Roth, Le~Lu, Jiamin Liu, Jianhua Yao, Ari Seff, Kevin Cherry, Lauren
  Kim, and Ronald~M Summers.
\newblock {Improving computer-aided detection using convolutional neural
  networks and random view aggregation}.
\newblock \emph{IEEE transactions on medical imaging}, 35\penalty0
  (5):\penalty0 1170--1181, 2016{\natexlab{b}}.

\bibitem[Simonovsky et~al.(2016)Simonovsky, Guti{\'{e}}rrez-Becker, Mateus,
  Navab, and Komodakis]{Simonovsky2016}
Martin Simonovsky, Benjam{\'{i}}n Guti{\'{e}}rrez-Becker, Diana Mateus, Nassir
  Navab, and Nikos Komodakis.
\newblock {A deep metric for multimodal registration}.
\newblock In \emph{International Conference on Medical Image Computing and
  Computer-Assisted Intervention}, pages 10--18. Springer, 2016.

\bibitem[Maes et~al.(1997)Maes, Collignon, Vandermeulen, Marchal, and
  Suetens]{Maes1997}
Frederik Maes, Andre Collignon, Dirk Vandermeulen, Guy Marchal, and Paul
  Suetens.
\newblock {Multimodality image registration by maximization of mutual
  information}.
\newblock \emph{Medical Imaging, IEEE Transactions on}, 16\penalty0
  (2):\penalty0 187--198, 1997.
\newblock ISSN 0278-0062.

\bibitem[Liao et~al.(2017)Liao, Miao, de~Tournemire, Grbic, Kamen, Mansi, and
  Comaniciu]{Liao2017}
Rui Liao, Shun Miao, Pierre de~Tournemire, Sasa Grbic, Ali Kamen, Tommaso
  Mansi, and Dorin Comaniciu.
\newblock {An Artificial Agent for Robust Image Registration.}
\newblock In \emph{AAAI}, pages 4168--4175, 2017.

\bibitem[Sokooti et~al.(2017)Sokooti, de~Vos, Berendsen, Lelieveldt, Isgum, and
  Staring]{Sokooti2017}
Hessam Sokooti, Bob de~Vos, Floris Berendsen, Boudewijn P.~F. Lelieveldt, Ivana
  Isgum, and Marius Staring.
\newblock {Nonrigid Image Registration Using Multi-scale 3D Convolutional
  Neural Networks}.
\newblock In \emph{Medical Image Computing and Computer Assisted Intervention},
  pages 232--239. 2017.
\newblock ISBN 978-3-319-66182-7.
\newblock \doi{10.1007/978-3-319-66182-7_27}.

\bibitem[Lv et~al.(2017)Lv, Yang, Zhang, and Wang]{Lv2017}
Jun Lv, Ming Yang, Jue Zhang, and Xiaoying Wang.
\newblock {Respiratory motion correction for free-breathing 3D abdominal MRI
  using CNN based image registration: a feasibility study}.
\newblock \emph{The British Journal of Radiology}, page 20170788, dec 2017.
\newblock ISSN 0007-1285.
\newblock \doi{10.1259/bjr.20170788}.

\bibitem[de~Vos et~al.(2017)de~Vos, Berendsen, Viergever, Staring, and
  I{\v{s}}gum]{DeVos2017}
Bob~D de~Vos, Floris~F Berendsen, Max~A Viergever, Marius Staring, and Ivana
  I{\v{s}}gum.
\newblock {End-to-End Unsupervised Deformable Image Registration with a
  Convolutional Neural Network BT - Deep Learning in Medical Image Analysis and
  Multimodal Learning for Clinical Decision Support : Third International
  Workshop, DLMIA 2017, and 7th International}.
\newblock pages 204--212. Springer International Publishing, Cham, 2017.
\newblock ISBN 978-3-319-67558-9.
\newblock \doi{10.1007/978-3-319-67558-9_24}.

\bibitem[Bahrami et~al.(2016)Bahrami, Shi, Zong, Shin, An, and
  Shen]{Bahrami2016}
Khosro Bahrami, Feng Shi, Xiaopeng Zong, Hae~Won Shin, Hongyu An, and Dinggang
  Shen.
\newblock {Reconstruction of 7T-Like Images from 3T MRI}.
\newblock \emph{IEEE Transactions on Medical Imaging}, 35\penalty0
  (9):\penalty0 2085--2097, 2016.
\newblock ISSN 1558254X.
\newblock \doi{10.1109/TMI.2016.2549918}.

\bibitem[Schlemper et~al.(2017)Schlemper, Caballero, Hajnal, Price, and
  Rueckert]{Schlemper2017}
J~Schlemper, J~Caballero, J~V Hajnal, A~Price, and D~Rueckert.
\newblock {A Deep Cascade of Convolutional Neural Networks for Dynamic MR Image
  Reconstruction}.
\newblock \emph{IEEE Transactions on Medical Imaging}, PP\penalty0
  (99):\penalty0 1, 2017.
\newblock ISSN 0278-0062 VO - PP.
\newblock \doi{10.1109/TMI.2017.2760978}.

\bibitem[Yang et~al.(2017{\natexlab{b}})Yang, Yu, Dong, Slabaugh, Dragotti, Ye,
  Liu, Arridge, Keegan, Guo, and Firmin]{Yang2017}
Guang Yang, Simiao Yu, Hao Dong, Greg Slabaugh, Pier~Luigi Dragotti, Xujiong
  Ye, Fangde Liu, Simon Arridge, Jennifer Keegan, Yike Guo, and David Firmin.
\newblock {DAGAN: Deep De-Aliasing Generative Adversarial Networks for Fast
  Compressed Sensing MRI Reconstruction}.
\newblock \emph{IEEE Transactions on Medical Imaging}, pages 1--1,
  2017{\natexlab{b}}.
\newblock ISSN 0278-0062.
\newblock \doi{10.1109/TMI.2017.2785879}.

\bibitem[Chartsias et~al.(2017)Chartsias, Joyce, Giuffrida, and
  Tsaftaris]{Chartsias2017}
Agisilaos Chartsias, Thomas Joyce, Mario~Valerio Giuffrida, and Sotirios~A.
  Tsaftaris.
\newblock {Multimodal MR Synthesis via Modality-Invariant Latent
  Representation}.
\newblock \emph{IEEE Transactions on Medical Imaging}, 0062\penalty0
  (JANUARY):\penalty0 1--1, 2017.
\newblock ISSN 0278-0062.
\newblock \doi{10.1109/TMI.2017.2764326}.

\bibitem[Chen et~al.(2017{\natexlab{b}})Chen, Zhang, Zhang, Liao, Li, Zhou, and
  Wang]{Chen2017}
Hu~Chen, Yi~Zhang, Weihua Zhang, Peixi Liao, Ke~Li, Jiliu Zhou, and Ge~Wang.
\newblock {Low-dose CT via convolutional neural network}.
\newblock \emph{Biomedical optics express}, 8\penalty0 (2):\penalty0 679--694,
  2017{\natexlab{b}}.
\newblock ISSN 2156-7085.

\bibitem[Kang et~al.(2017)Kang, Min, and Ye]{Kang2017}
Eunhee Kang, Junhong Min, and Jong~Chul Ye.
\newblock {A deep convolutional neural network using directional wavelets for
  low-dose X-ray CT reconstruction}.
\newblock \emph{Medical Physics}, 44\penalty0 (10):\penalty0 e360--e375, oct
  2017.
\newblock ISSN 00942405.
\newblock \doi{10.1002/mp.12344}.

\bibitem[Jin et~al.(2017)Jin, McCann, Froustey, and Unser]{Jin2017}
Kyong~Hwan Jin, Michael~T. McCann, Emmanuel Froustey, and Michael Unser.
\newblock {Deep Convolutional Neural Network for Inverse Problems in Imaging}.
\newblock \emph{IEEE Transactions on Image Processing}, 26\penalty0
  (9):\penalty0 4509--4522, 2017.
\newblock ISSN 10577149.
\newblock \doi{10.1109/TIP.2017.2713099}.

\bibitem[Nie et~al.(2016)Nie, Cao, Gao, Wang, and Shen]{Nie2016}
Dong Nie, Xiaohuan Cao, Yaozong Gao, Li~Wang, and Dinggang Shen.
\newblock {Estimating CT Image from MRI Data Using 3D Fully Convolutional
  Networks BT - Deep Learning and Data Labeling for Medical Applications: First
  International Workshop, LABELS 2016, and Second International Workshop, DLMIA
  2016, Held in Conjunction with MICC}.
\newblock pages 170--178. Springer International Publishing, Cham, 2016.
\newblock ISBN 978-3-319-46976-8.
\newblock \doi{10.1007/978-3-319-46976-8_18}.

\bibitem[Han(2017)]{Han2017}
Xiao Han.
\newblock {MR-based synthetic CT generation using a deep convolutional neural
  network method}.
\newblock \emph{Medical Physics}, 44\penalty0 (4):\penalty0 1408--1419, apr
  2017.
\newblock ISSN 00942405.
\newblock \doi{10.1002/mp.12155}.

\bibitem[Li et~al.(2014)Li, Zhang, Suk, Wang, Li, Shen, and Ji]{Li2014}
Rongjian Li, Wenlu Zhang, Heung-Il Suk, Li~Wang, Jiang Li, Dinggang Shen, and
  Shuiwang Ji.
\newblock {Deep Learning Based Imaging Data Completion for Improved Brain
  Disease Diagnosis}.
\newblock \emph{Medical image computing and computer-assisted intervention :
  MICCAI ... International Conference on Medical Image Computing and
  Computer-Assisted Intervention}, 17\penalty0 (0 3):\penalty0 305--312, 2014.

\bibitem[Oktay et~al.(2016)Oktay, Bai, Lee, Guerrero, Kamnitsas, Caballero,
  de~Marvao, Cook, O'Regan, and Rueckert]{Oktay2016}
Ozan Oktay, Wenjia Bai, Matthew Lee, Ricardo Guerrero, Konstantinos Kamnitsas,
  Jose Caballero, Antonio de~Marvao, Stuart Cook, Declan O'Regan, and Daniel
  Rueckert.
\newblock {Multi-input Cardiac Image Super-Resolution Using Convolutional
  Neural Networks}.
\newblock In \emph{Medical Image Computing and Computer-Assisted Intervention
  -- MICCAI 2016: 19th International Conference, Athens, Greece, October 17-21,
  2016, Proceedings, Part III}, pages 246--254, 2016.
\newblock ISBN 3642236286.
\newblock \doi{10.1007/978-3-319-46726-9_29}.

\bibitem[Umehara et~al.(2017)Umehara, Ota, and Ishida]{Umehara2017}
Kensuke Umehara, Junko Ota, and Takayuki Ishida.
\newblock {Application of Super-Resolution Convolutional Neural Network for
  Enhancing Image Resolution in Chest CT}, oct 2017.
\newblock ISSN 1618727X.

\bibitem[Benou et~al.(2017)Benou, Veksler, Friedman, and {Riklin
  Raviv}]{Benou2017}
A~Benou, R~Veksler, A~Friedman, and T~{Riklin Raviv}.
\newblock {Ensemble of expert deep neural networks for spatio-temporal
  denoising of contrast-enhanced MRI sequences}.
\newblock \emph{Medical Image Analysis}, 42:\penalty0 145--159, 2017.
\newblock \doi{10.1016/j.media.2017.07.006}.

\bibitem[Jiang et~al.(2017)Jiang, Dou, Vosters, Xu, Sun, and Tan]{Jiang2017}
Dongsheng Jiang, Weiqiang Dou, Luc Vosters, Xiayu Xu, Yue Sun, and Tao Tan.
\newblock {Denoising of 3D magnetic resonance images with multi-channel
  residual learning of convolutional neural network}.
\newblock \emph{arxiv.org}, 2017.

\bibitem[Yang et~al.(2017{\natexlab{c}})Yang, Chen, Liu, Zhong, Qin, Medical,
  and 2017]{Yang2017a}
W~Yang, Y~Chen, Y~Liu, L~Zhong, G~Qin, Z~Lu Medical, and Undefined 2017.
\newblock {Cascade of multi-scale convolutional neural networks for bone
  suppression of chest radiographs in gradient domain}.
\newblock \emph{Medical Image Analysis}, 35:\penalty0 421--433,
  2017{\natexlab{c}}.

\bibitem[Zhang et~al.(2017{\natexlab{b}})Zhang, Chu, and Yu]{Zhang2017}
Yanbo Zhang, Ying Chu, and Hengyong Yu.
\newblock {Reduction of metal artifacts in x-ray CT images using a
  convolutional neural network}.
\newblock volume 10391, pages 103910V--10391--11, 2017{\natexlab{b}}.

\bibitem[Qayyum et~al.(2017)Qayyum, Anwar, Awais, and Majid]{Qayyum2017}
Adnan Qayyum, Syed~Muhammad Anwar, Muhammad Awais, and Muhammad Majid.
\newblock {Medical image retrieval using deep convolutional neural network}.
\newblock \emph{Neurocomputing}, 266\penalty0 (Supplement C):\penalty0 8--20,
  2017.
\newblock ISSN 0925-2312.
\newblock \doi{https://doi.org/10.1016/j.neucom.2017.05.025}.

\bibitem[Yan et~al.(2017)Yan, Wang, Lu, Zhang, Harrison, Bagheri, and
  Summers]{Yan2017}
Ke~Yan, Xiaosong Wang, Le~Lu, Ling Zhang, Adam Harrison, Mohammadhad Bagheri,
  and Ronald Summers.
\newblock Deep lesion graphs in the wild: Relationship learning and
  organization of significant radiology image findings in a diverse large-scale
  lesion database.
\newblock \emph{arXiv preprint arXiv:1711.10535}, 2017.

\bibitem[Chow and Paramesran(2016)]{Chow2016}
Li~Sze Chow and Raveendran Paramesran.
\newblock {Review of medical image quality assessment}.
\newblock \emph{Biomedical Signal Processing and Control}, 27:\penalty0
  145--154, may 2016.
\newblock ISSN 1746-8094.
\newblock \doi{10.1016/J.BSPC.2016.02.006}.

\bibitem[Wu et~al.(2017)Wu, Cheng, Li, Lei, and Wang]{Wu2017}
L~Wu, JZ~Cheng, S~Li, B~Lei, and T~Wang.
\newblock {FUIQA: Fetal ultrasound image quality assessment with deep
  convolutional networks}.
\newblock \emph{IEEE Trans Cybernetics}, 45\penalty0 (5):\penalty0 1336 --
  1349, 2017.

\bibitem[Abdi et~al.(2017)Abdi, Luong, Tsang, and Imaging]{Abdi2017}
AH~Abdi, C~Luong, T~Tsang, and G~Allan Imaging.
\newblock {Automatic Quality Assessment of Echocardiograms Using Convolutional
  Neural Networks: Feasibility on the Apical Four-Chamber View}.
\newblock \emph{IEEE Trans Med Imaging}, 36\penalty0 (6):\penalty0 1221--1230,
  2017.

\bibitem[Esses et~al.(2017)Esses, Lu, Zhao, Shanbhogue, Dane, Bruno, and
  Chandarana]{Esses2017}
Steven~J. Esses, Xiaoguang Lu, Tiejun Zhao, Krishna Shanbhogue, Bari Dane, Mary
  Bruno, and Hersh Chandarana.
\newblock {Automated image quality evaluation of T2-weighted liver MRI
  utilizing deep learning architecture}.
\newblock \emph{Journal of Magnetic Resonance Imaging}, jun 2017.
\newblock ISSN 10531807.
\newblock \doi{10.1002/jmri.25779}.

\bibitem[Gale et~al.(2017)Gale, Oakden-Rayner, Carneiro, Bradley, and
  Palmer]{Gale2017}
William Gale, Luke Oakden-Rayner, Gustavo Carneiro, Andrew~P Bradley, and
  Lyle~J Palmer.
\newblock {Detecting hip fractures with radiologist-level performance using
  deep neural networks}.
\newblock \emph{arXiv preprint arXiv:1711.06504}, 2017.

\bibitem[Grewal et~al.(2017)Grewal, Srivastava, Kumar, and
  Varadarajan]{Grewal2017}
Monika Grewal, Muktabh~Mayank Srivastava, Pulkit Kumar, and Srikrishna
  Varadarajan.
\newblock {RADNET: Radiologist Level Accuracy using Deep Learning for
  HEMORRHAGE detection in CT Scans}.
\newblock \emph{arXiv preprint arXiv:1710.04934}, 2017.

\bibitem[Jamaludin et~al.(2017)Jamaludin, Lootus, Kadir, Zisserman, Urban,
  Batti{\'{e}}, Fairbank, and McCall]{Jamaludin2017}
Amir Jamaludin, Meelis Lootus, Timor Kadir, Andrew Zisserman, Jill Urban,
  Michele~C Batti{\'{e}}, Jeremy Fairbank, and Iain McCall.
\newblock {Automation of reading of radiological features from magnetic
  resonance images (MRIs) of the lumbar spine without human intervention is
  comparable with an expert radiologist}.
\newblock \emph{European Spine Journal}, 26\penalty0 (5):\penalty0 1374--1383,
  2017.
\newblock ISSN 1432-0932.
\newblock \doi{10.1007/s00586-017-4956-3}.

\bibitem[Kooi et~al.(2017)Kooi, Litjens, van Ginneken, Gubern-M{\'{e}}rida,
  S{\'{a}}nchez, Mann, den Heeten, and Karssemeijer]{Kooi2017}
Thijs Kooi, Geert Litjens, Bram van Ginneken, Albert Gubern-M{\'{e}}rida,
  Clara~I S{\'{a}}nchez, Ritse Mann, Ard den Heeten, and Nico Karssemeijer.
\newblock {Large scale deep learning for computer aided detection of
  mammographic lesions}.
\newblock \emph{Medical image analysis}, 35:\penalty0 303--312, 2017.
\newblock ISSN 1361-8415.

\bibitem[Larson et~al.(2017)Larson, Chen, Lungren, Halabi, Stence, and
  Langlotz]{Larson2017}
David~B Larson, Matthew~C Chen, Matthew~P Lungren, Safwan~S Halabi, Nicholas~V
  Stence, and Curtis~P Langlotz.
\newblock {Performance of a Deep-Learning Neural Network Model in Assessing
  Skeletal Maturity on Pediatric Hand Radiographs}.
\newblock \emph{Radiology}, page 170236, nov 2017.
\newblock ISSN 0033-8419.
\newblock \doi{10.1148/radiol.2017170236}.

\bibitem[Merkow et~al.(2017)Merkow, Luftkin, Nguyen, Soatto, Tu, and
  Vedaldi]{Merkow2017}
Jameson Merkow, Robert Luftkin, Kim Nguyen, Stefano Soatto, Zhuowen Tu, and
  Andrea Vedaldi.
\newblock {DeepRadiologyNet: Radiologist Level Pathology Detection in CT Head
  Images}.
\newblock \emph{arXiv preprint arXiv:1711.09313}, 2017.

\bibitem[Olczak et~al.(2017)Olczak, Fahlberg, Maki, Razavian, Jilert, Stark,
  Sk{\"{o}}ldenberg, and Gordon]{Olczak2017}
Jakub Olczak, Niklas Fahlberg, Atsuto Maki, Ali~Sharif Razavian, Anthony
  Jilert, Andr{\'{e}} Stark, Olof Sk{\"{o}}ldenberg, and Max Gordon.
\newblock {Artificial intelligence for analyzing orthopedic trauma
  radiographs}.
\newblock \emph{Acta Orthopaedica}, 88\penalty0 (6):\penalty0 581--586, nov
  2017.
\newblock ISSN 1745-3674.
\newblock \doi{10.1080/17453674.2017.1344459}.

\end{thebibliography}

\end{document}